\documentclass[10pt,journal,letterpaper,compsoc]{IEEEtran}

\usepackage{latexsym,bm,amsmath,amssymb}
\usepackage{graphicx}
\usepackage{graphics}

\usepackage{subfigure}
\usepackage{multirow}
\usepackage{float}
\usepackage{algorithm}
\usepackage{algorithmicx}
\usepackage{algpseudocode}
\usepackage{array}
\usepackage{xspace}

\makeatletter
\DeclareRobustCommand\onedot{\futurelet\@let@token\@onedot}
\def\@onedot{\ifx\@let@token.\else.\null\fi\xspace}

\def\eg{\emph{e.g}\onedot} 
\def\ie{\emph{i.e}\onedot}

\makeatother

\newcommand{\DoubleFigureWidth}{230pt}
\newcommand{\QuaterFigureWidth}{120pt}

\begin{document}

\title{EFANNA : An Extremely Fast Approximate Nearest Neighbor Search Algorithm Based on kNN Graph}%

\author{Cong Fu, \ \ Deng Cai
	\thanks{C. Fu and D. Cai are with the State Key Lab of CAD\&CG, College of Computer Science,
		Zhejiang University, Hangzhou, Zhejiang, China, 310058.
		Email: \texttt{15267003518@163.com, dengcai@cad.zju.edu.cn}.}
}

\IEEEcompsoctitleabstractindextext{%
	\vspace{6mm}
\begin{abstract} 
	Approximate nearest neighbor (ANN) search is a fundamental problem in many areas of data mining, machine learning and computer vision. The performance of traditional hierarchical structure (tree) based methods decreases as the dimensionality of data grows, while hashing based methods usually lack efficiency in practice. Recently, the graph based methods have drawn considerable attention. The main idea is that \emph{a neighbor of a neighbor is also likely to be a neighbor}, which we refer as \emph{NN-expansion}. These methods construct a $k$-nearest neighbor ($k$NN) graph offline. And at online search stage, these methods find candidate neighbors of a query point in some way (\eg, random selection), and then check the neighbors of these candidate neighbors for closer ones iteratively. Despite some promising results, there are mainly two problems with these approaches: 1) These approaches tend to converge to local optima. 2) Constructing a $k$NN graph is time consuming. We find that these two problems can be nicely solved when we provide a good initialization for NN-expansion. In this paper, we propose EFANNA, an extremely fast approximate nearest neighbor search algorithm based on $k$NN Graph. Efanna nicely combines the advantages of hierarchical structure based methods and  nearest-neighbor-graph based methods. Extensive experiments have shown that EFANNA outperforms the state-of-art algorithms both on approximate nearest neighbor search and approximate nearest neighbor graph construction. To the best of our knowledge, EFANNA is the fastest algorithm so far both on approximate nearest neighbor graph construction and approximate nearest neighbor search. A library EFANNA based on this research is released on Github.

\end{abstract}

}

\maketitle
\IEEEdisplaynotcompsoctitleabstractindextext
\IEEEpeerreviewmaketitle

\section{Introduction}

\IEEEPARstart Nearest neighbor search plays an important role in many applications of data mining, machine learning and computer vision. When dealing with sparse data (\eg, document retrieval), one can use advanced index structures (\eg, inverted index) to solve this problem. However, for data with dense features, the cost for finding the exact nearest neighbor is $O(N)$, where $N$ is the number of points in the database. It's very time consuming when the data set is large. So people turn to Approximate Nearest neighbor (ANN) search in practice \cite{Arya1998Optimal,Nitin2010Survey}. Many work has been done to carry out the ANN search with high accuracy but low computational complexity.

There are mainly two types of methods in ANN search. The first type methods are hierarchical structure (tree) based methods, such as KD-tree \cite{Bentley1975Multidimensional,Friedman1975An} ,Randomized KD-tree \cite{Silpaanan2008Optimised}, K-means tree \cite{Fukunaga1975A}. These methods perform very well when the dimension of the data is relatively low. However, the performance decreases dramatically as the dimension of the data increases \cite{Silpaanan2008Optimised}. The second type methods are hashing based methods, such as Locality Sensitive Hashing (LSH) \cite{Gionis1999Similarity}, Spectral Hashing \cite{Weiss2008Spectral},  Iterative Quantization \cite{Yunchao2011Iterative} and so on. Please see \cite{WangSSJ14} for a detailed survey on various hashing methods. These methods generate
binary codes for high dimensional real vectors while try to preserve the similarity among original real vectors. Thus, all the real vectors fall into different hashing buckets (with different binary codes). Ideally, if neighbor vectors fall into the same bucket or the nearby buckets (measured by the hamming distance of two binary codes), the hashing based methods can efficiently retrieve the nearest neighbors of a query point. However, there is no guarantee that all the neighbor vectors will fall into the nearby buckets. To ensure the high \emph{recall} (the number of true neighbors within the returned points set divides by the number of required neighbors), one needs to examine many hashing buckets (\ie, enlarge the search radius in hamming space), which results a high computational complexity. Please see \cite{Jin2014Fast} for a detailed analysis.

Recently, graph based methods have drawn considerable attention \cite{Kgraph2014,Hajebi2011Fast,Jin2014Fast}. The essential idea behind these approaches is that \emph{a neighbor of a neighbor is also likely to be a neighbor}, which we refer as \emph{NN-expansion}. These methods construct a $k$-nearest neighbor ($k$NN) graph offline. And at online search stage, these methods find the candidate neighbors of a query point in some way (\eg, random selection \cite{Kgraph2014,Hajebi2011Fast}), and then check the neighbors of these candidate neighbors for closer ones iteratively. One problem of this approach is that the NN-expansion is easily to converge to local optima and result in a low recall. \cite{Jin2014Fast} tries to solve this problem  by providing better initialization for a query point. Instead of random selection, \cite{Jin2014Fast} uses hashing based methods for initialization. This approach is named as Iterative Expanding Hashing (IEH) and achieves significant better results than the corresponding hashing based methods.

Another challenge on using graph based methods is the high computational cost in building the $k$NN graph, especially when the database is large. There are many efforts that have been put into reducing the time complexity of $k$NN graph construction. \cite{Bentley1980Multidimensional,Clarkson1983Fast,Pravin1989AnO,Paredes2006Practical} try to speed up an exact $k$NN graph construction. However, these approaches are still not efficient enough in the context of big data. Instead of building an exact $k$NN graph, recent researchers try to build an approximated $k$NN graph efficiently. The \emph{NN-expansion} idea again can be used to build an approximated $k$NN graph. \cite{Dong2011Efficient} proposed NN-descent to efficiently build an approximate $k$NN graph. \cite{Chen2009Fast,Gan2012Scalable,Zhang2013Fast,tang2016visualizing} try to build an approximate $k$NN graph in a divide-and-conquer manner. Their algorithms mainly contain three phrases. Firstly, they divide the whole data set into small subsets multiple times. Secondly, they do brute force search within the subsets and get lots of overlapping subgraphs. Finally, they merge the subgraphs and refine the graph with techniques similar to NN-expansion. 
Although an approximated $k$NN graph can be efficiently constructed, there are no formal study on how the performance of graph based search methods will be affected if one uses an approximated $k$NN graph instead of an exact $k$NN graph.

To tackle above problems, we propose a novel graph-based  approximate nearest neighbor search framework EFANNA in this paper. EFANNA is an abbreviation for two algorithms: {\bf E}xtremely {\bf F}ast {\bf A}pproximate $k$-{\bf N}earest {\bf N}eighbor graph construction {\bf A}lgorithm and {\bf E}xtremely {\bf F}ast {\bf A}pproximate {\bf N}earest {\bf N}eighbor search {\bf A}lgorithm based on $k$NN graph. Our algorithm is based on a simple observation: the performance of both NN-expansion and NN-descent is very sensitive with the initialization. 

Our EFANNA index contains two parts: the multiple randomized hierarchical structures (\eg, randomized truncated KD-tree) and an approximate $k$-nearest neighbor graph. 

At offline stage, EFANNA divides the data set multiple times into a number of subsets in a fast and hierarchical way, producing multiple randomized hierarchical structures. Then EFANNA constructs an approximate $k$NN graph by conquering bottom-up along the structures. When conquering, EFANNA takes advantage of the structures to locate the closest possible neighbors, and use these candidates to update the graph, which reduces the computation cost than using all the points in the subtree. Finally we refine the graph similar to NN-descent \cite{Dong2011Efficient}, which is based on the NN-expansion idea and optimized with techniques like local join, sampling, and early termination \cite{Dong2011Efficient}. 

At online search stage, EFANNA first search in the hierarchical structures to get candidate neighbors for a given query. Then EFANNA refines the results using NN-expansion on the approximate $k$NN graph. Extensive experimental results show that our approach outperforms the the-state-of-the-art approximate nearest neighbor search algorithms significantly.

It is worthwhile to highlight the contributions of our paper as follows:
\begin{itemize}
	\item EFANNA outperforms state-of-the-art ANN search algorithms significantly. Particularly, EFANNA outperforms Flann \cite{Muja2014Scalable}, one of the most popular ANN search library, in index size, search speed and search accuracy. 
	\item EFANNA can build  an approximate $k$NN graph with hundreds times speed-up over brute-force graph building on million scale datasets. Considering many unsupervised and semi-supervised machine learning algorithms \cite{MR_JMLR_06,shi2000normalized} are based on a nearest neighbor graph. EFANNA provides the possibility to examine the effectiveness of all these algorithms on large scale datasets.
	\item We show by experimental results that with an approximate $k$NN graph of low accuracy constructed by EFANNA, graph-based ANN search methods (\eg, EFANNA) still perform very good. This is because the "error" neighbors of approximate $k$NN graph constructed by EFANNA are actually neighbors a little farther. This property is never explored in the previous work.
\end{itemize}

The remainder of this paper is organized as follows. In Section \ref{sec2} we will introduce some related work. Our EFANNA algorithm is presented in section \ref{sec3}. In section \ref{sec4}, we will report the experimental results and show the performance of EFANNA comprehensively. In section \ref{sec5} we will talk about our open library and in section \ref{sec6} we will draw a conclusion.

\section{Related work}
\label{sec2}

Nearest neighbor search \cite{yianilos1993data} has been a hot topic during the last decades. Due to the intrinsic difficulty of exact nearest neighbor search, the approximate nearest neighbor (ANN) search algorithms \cite{Arya1998Optimal,indyk1998approximate} are widely studied and the researchers expect to sacrifice a little searching accuracy to lower the time cost as much as possible. 

Hierarchical index based (tree based) algorithms, such as KD-tree \cite{Friedman1975An}, have gained early success on approximate nearest neighbor search problems. However, it's proved to be inefficient when the dimensionality of data grows high. Many new hierarchical structure based methods \cite{Silpaanan2008Optimised,Brin1996Near,Nister2006Scalable} are presented to address this limitation. Randomized KD-tree \cite{Silpaanan2008Optimised} and Kmeans tree \cite{Nister2006Scalable} are absorbed into a well-known open source library FLANN \cite{Muja2009Fast}, which has gained wide popularity.

Hashing based algorithms \cite{Gionis1999Similarity,Weiss2008Spectral} aim at finding proper ways to generate binary codes for data points and preserve their similarity in original feature space. These methods can be treated as dividing the data space with multiple hyperplanes and representing each resulting polyhedron with a binary code. Learning the hashing functions with different constraints will result in different partition of data space. One of the most famous algorithms is Locality Sensitive Hashing (LSH) \cite{Gionis1999Similarity}, which is essentially based on random projection \cite{bingham2001random}. Many other variants \cite{Weiss2008Spectral, Yunchao2011Iterative,Heo2012Spherical,Jin2013Complementary} are proposed based on different constraints. And the constraints reflect what they think is the proper way to partition the data space. 

Both the hashing based methods and tree based methods have the same goal. They expect to put neighbors into the same hashing bucket (or node). However, there is no theoretical guarantee of this expectation. To increase the search \emph{recall} (the number of true neighbors within the returned points divides by the number of required neighbors), one needs to check the ``nearby'' buckets or nodes. With high dimensional data, one polyhedron may have a large amount of neighbor polyhedrons, (for example, a bucket with 32 bit hashing code has 32 neighbor buckets with 1 hamming radius distance), which makes locating true neighbors hard \cite{Jin2014Fast}. 

Recently graph based techniques have drawn considerable attention \cite{Kgraph2014,Hajebi2011Fast,Jin2014Fast}. The main idea of these methods is \emph{a neighbor of a neighbor is also likely to be a neighbor}, which we refer as \emph{NN-expansion}. At offline stage, they need to build a $k$NN graph, which can be regraded as a big table recording the top $k$ closest neighbors of each point in database. At online stage, given a query point, they first assign the query some points as initial candidate neighbors, and then check the neighbors of the neighbors iteratively to locate closer neighbors. 
Graph Nearest neighbor Search (GNNS) \cite{Hajebi2011Fast} randomly generate the initial candidate neighbors while Iterative Expanding Hashing (IEH) \cite{Jin2014Fast} uses hashing algorithms to generate the initial candidate neighbors. 

Since all the graph based methods need a  $k$NN graph as the index structure, how to build a $k$NN graph efficiently became a crucial problem, especially when the database is large. Many work has been done on building either exact or approximate $k$NN graph. \cite{Bentley1980Multidimensional,Clarkson1983Fast,Paredes2006Practical,Pravin1989AnO} try to build an exact $k$NN graph quickly. However, these approaches are still not efficient enough in the context of big data. 
Recently, researchers try to build an approximated $k$NN graph efficiently. Again, the NN-expansion idea can be used here.
\cite{Dong2011Efficient} proposed an \emph{NN-descent} algorithm to efficiently build an approximate $k$NN graph. The basic idea of NN-descent is similar to NN-expansion but the details are different. NN-descent uses many techniques (\eg, \textbf{Local join} and \textbf{Sampling}) to efficiently refine the graph. Please see \cite{Dong2011Efficient} for details.  

Instead of initializing the $k$NN graph randomly, \cite{Chen2009Fast,Gan2012Scalable,Zhang2013Fast,tang2016visualizing} uses some divide-and-conquer methods. Their initialization contains two parts. Firstly, they divide the whole data set into small subsets multiple times. Secondly, they do brute force search within the subsets and get lots of overlapping subgraphs. These subgraphs can be merged together to serve as the initialization of $k$NN graph. The NN-expansion like techniques can then be used to refine the graph.
The division step of \cite{Chen2009Fast} is based on a spectral bisection and they proposed two different versions, overlap and glue division. \cite{Zhang2013Fast} use Anchor Graph Hashing \cite{liu2011hashing} to produce the division. \cite{Gan2012Scalable} uses recursive random division, dividing  orthogonal to the principle direction of randomly sampled data in subsets. \cite{tang2016visualizing} uses random projection trees to partition the datasets.

\begin{algorithm}[t]
	\caption{EFANNA Search Algorithm}
	\label{alg:search}
	\begin{algorithmic}[1]
		\Require data set D, query vector $q$, the number $K$ of required nearest neighbors, EFANNA index (including tree set $S_{tree}$ and $k$NN graph $G$), the candidate pool size $P$, the expansion factor $E$, the iteration number $I$.
		\Ensure approximate nearest neighbor set $ANNS$ of the query 
		\State $iter = 0$
		\State $NodeList = \emptyset$
		\State candidate set $C = \emptyset$
		\State suppose the maximal number of points of leaf node is $S_{leaf}$
		\State suppose the number of trees is $N_{tree}$
		\State then the maximal node check number is $N_{node} = P \div S_{leaf} \div N_{tree} + 1$ 
		\ForAll{tree $i$ in $S_{tree}$}
			\State Depth-first search $i$ for top $N_{node}$ closest leaf nodes according to respective tree search criteria, add to $NodeList$
		\EndFor
		\State add the points belonging to the nodes in $NodeList$ to $C$
		\State keep $E$ points in $C$ which are closest to $q$. 
		\While{$iter < I$}
		\State candidate set $CC = \emptyset$		
		\ForAll{point $n$ in $C$}
		\State $S_{n}$ is the neighbors of point $n$ based on $G$.
		\ForAll{point $nn$ in $S_{n}$}
		\If {$nn$ hasn't been checked}
		\State put $nn$ into $CC$.
		\EndIf
		\EndFor
		\EndFor
		\State move all the points in $CC$ to $C$ and keep $P$ points in $C$ which are closest to $q$.
		\State $iter = iter + 1$
		\EndWhile
		\State return $ANNS$ as the closet $K$ points to $q$ in $C$.
	\end{algorithmic}
\end{algorithm}

\cite{Zhang2013Fast} and \cite{tang2016visualizing} claim to outperform NN-descent \cite{Dong2011Efficient} significantly. However, based on their reported results and our analysis, there seems a misunderstanding of NN-descent \cite{Dong2011Efficient}. Actually, NN-descent is quite different than NN-expansion. The method compared in \cite{Zhang2013Fast} and \cite{tang2016visualizing} should be NN-expansion instead of NN-descent. Please see Section \ref{sec:NNexpNNdes} for details.

\section{EFANNA Algorithms for ANN search}\label{sec3}

We will introduce our EFANNA algorithms in this section. EFANNA algorithms include offline index building part and online ANN search algorithm. The EFANNA index contains two parts: multiple hierarchical structures (\eg, randomized truncated KD-tree) and an approximate $k$NN graph. We will first show how to use EFANNA index to carry out online ANN search. Then we will show how to build the EFANNA index in a divide-conquer-refinement manner. 

\begin{algorithm}[t]
	\caption{EFANNA Tree Building Algorithm}
	\label{alg:treeBuild}
	\begin{algorithmic}[1]
		\Require the data set $D$, the number of trees $T$, the number of points in a leaf node $K$.
		\Ensure the randomized truncated KD-tree set $S$
		\\
		\Function {BuildTree}{$Node, PointSet$}
		\If{size of PointSet $< K$}
		\State \Return
		\Else
		\State Randomly choose dimension $d$.
		\State Calculate the mean $mid$ over $PointSet$ on dimension $d$.
		\State Divide $PointSet$ evenly into two subsets, $LeftHalf$ and  $RightHalf$, according to $mid$.
		\State \Call{BuildTree}{$Node.LeftChild, LeftHalf$}
		\State \Call{BuildTree}{$Node.RightChild, RightHalf$}
		\EndIf
		\State \Return
		\EndFunction
		\State
		\ForAll{$i = 1$ to $T$}
		\State \Call{BuildTree}{$Root_i, D$}
		\State Add $Root_i$ to $S$.
		\EndFor
	\end{algorithmic}
\end{algorithm}

\subsection{ANN search with EFANNA index}

EFANNA is a graph based method. The main idea is providing better initialization for NN-expansion to improve the performance significantly. The multiple hierarchical structures is used for initialization and the approximate $k$NN graph is used for NN-expansion.  

There are many possible hierarchical structures (\eg hierarchical clustering \cite{Muja2014Scalable} or randomized division tree \cite{dasgupta2008random}) can be used in our index structure. In this paper, we only report the results using randomized \emph{truncated} KD-tree. The details on the difference between this structure and the traditional randomized KD-tree \cite{Silpaanan2008Optimised} will be discussed in the next subsection. Based on this randomized truncated KD-tree, we can get the initial neighbor candidates given a query $q$. We then refine the result with NN-expansion, \ie, we check the neighbors of $q$'s neighbors according to the approximate $k$NN graph to get closer neighbors. Algorithm \ref{alg:search} shows the detailed procedure.

There are three essential parameters in our ANN search algorithm: the expansion factor $E$, the  the candidate pool size $P$ and the iteration number $I$. In our experiment, we found $I = 4$ is enough and thus we fixed $I = 4$. The trade-off between search speed and accuracy can be made through tuning parameters $E$ and $P$. In other words, larger $E$ and larger $P$ sacrifice the search speed for high accuracy.

\begin{algorithm}[t]
	\caption{Hierarchical Divide-and-Conquer Algorithm ($k$NN Graph Initialization)}
	\label{alg:generalBuild}
	\begin{algorithmic}[1]
		\Require the data set $D$, the $k$ in approximate $k$NN graph , the randomized truncated KD-tree set $S$ built with Algorithm \ref{alg:treeBuild}, the conquer-to depth $Dep$.
		\Ensure approximate $k$NN graph $G$.
		\State \%\%Division step
		\State Using Algorithm \ref{alg:treeBuild} to build tree, which leads to the input $S$
		\State
		
		\State \%\% Conquer step
		\State $G = \emptyset$
		\ForAll{point $i$ in $D$}
		\State Candidate pool $C = \emptyset$
		\ForAll{binary tree $t$ in $S$}
		\State search in tree $t$ with point $i$ to the leaf node. \State add all the point in the leaf node to $C$.
		\State $d = depth\ of\ the\ leaf\ node$
		\While{$d > Dep$}
		\State $d = d -1$
		\State  Depth-first-search in the tree $t$ with point $i$ to depth $d$. Suppose $N$ is the non-leaf node on the search path with depth $d$. Suppose $Sib$ is the child node of $N$. And $Sib$ is not on the search path of point $i$.
		\State  Depth-first-search to the leaf node in the subtree of $Sib$ with point $i$ . Add all the points in the leaf node to $C$.
		\EndWhile
		\EndFor
		\State Reserve $K$ closest points to $i$ in $C$.
		\State 
		\State Add $C$ to $G$.
		\EndFor
		
	\end{algorithmic}
\end{algorithm}

\subsection{EFANNA Index Building Algorithms \uppercase\expandafter{\romannumeral1} : Tree Buidling}
\label{treebuildalg}

One part of the EFANNA index is a multiple hierarchical structures. There are many possible hierarchical structures (\eg hierarchical clustering \cite{Muja2014Scalable} or randomized division tree \cite{dasgupta2008random}) can be used. In this paper, we only report the results using randomized \emph{truncated} KD-tree. Please see Algorithm \ref{alg:treeBuild} for details on building randomized truncated KD-trees. 

The only difference between randomized \emph{truncated} KD-tree and the traditional 
randomized KD-tree is that leaf node in our trees has $K$ ($K=10$ in our experiments) points instead of 1. This change makes the tree building in EFANNA much faster than the the traditional 
randomized KD-tree. Please see Table \ref{IndexTimeTable} in the experiments for details.

The randomized truncated KD-tree built in this step is used not only in the on-line search stage, but also in the approximate $k$NN graph construction stage. See the next section for details.

\subsection{EFANNA Index Building Algorithms \uppercase\expandafter{\romannumeral2} : Approximate $k$NN Graph Construction} 

Another part of the EFANNA index is an approximate $k$NN graph. We use the similar methodology to efficiently build the approximate $k$NN graph as with ANN search. It contains two stage. At first stage, we regard the trees built in the previous part as multiple overlapping divisions over the data set, and we perform the conquering step along the tree structures to get an initial $k$NN graph. At second stage, we use NN-descent \cite{Dong2011Efficient} to refine the $k$NN graph.

\begin{figure}
	\centering
	\includegraphics[width=\DoubleFigureWidth]{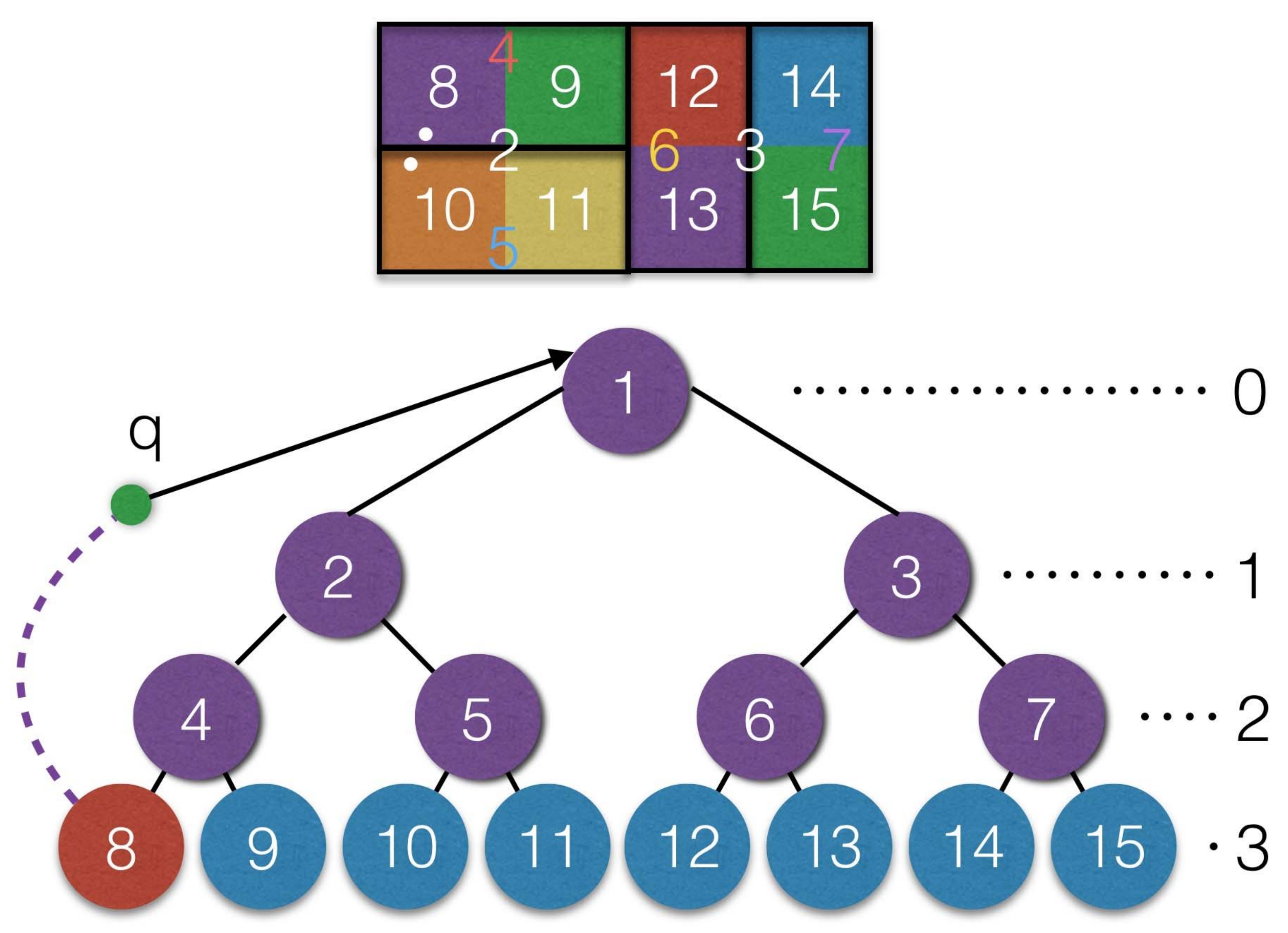}
	\caption{An example for our hierarchical divide-and-conquer algorithm.}
	\label{DivideConquerFig}
\end{figure}

\subsubsection{Hierarchical Randomized Divide-and-Conquer}

\cite{Dong2011Efficient} uses a random $k$NN graph as the initialization and refined it with the NN-descent algorithm to get a $k$NN graph with high accuracy. Our idea is very simple. We try to provide a better initialization for NN-descent. A good initialization should produce an initial $k$NN graph with certain accuracy within short time. 

Divide-and-conquer is a good strategy to achieve this goal. A normal divide-and-conquer process first breaks the problem into subproblems recursively until the subproblem is small and easy enough to solve. Then the solutions of subproblems are combined to get a solution to the original problem. For approximate graph construction, the division part is easy. We can divide the data set as the way the tree was constructed (Section \ref{treebuildalg}). Then we merge sibling nodes recursively upwards from leaf. With only one tree, we need to conquer to root to get a full connected graph. But if we directly conquer from leaf to root, the computational cost grows exponentially upwards and is no less than brute-force graph construction. 



To reduce the computational complexity, we need to find a better conquer strategy and avoid conquering to root. We follow the inspiration of previous work \cite{Gan2012Scalable}, which said multiple randomized division can produce overlapping subsets. As a result, the conquer doesn't need to be carried out to root. In addition, our motivation on better conquer strategy is to reduce the number of points involved in conquering at each level and keep points'  ``quality'' (we make sure that we always choose the closest possible points for conquer). For example, in Fig \ref{DivideConquerFig}, if we know that point $q$ in node (or subset) $8$ is closer to the area of node $10$, then we just need to consider the points in node $10$ when conquering $4$ and $5$ at level $1$ with $q$.

In other words, when we conquer two sibling non-leaf nodes (sibling leaf nodes can be conquered directly), for some points in one subtree, we may just consider the ``closest'' possible leaf node in the sibling subtree. Because the rest leaf nodes are farther, the points in them are also likely to be farther, thus excluded from distance calculating. We can regard the tree as a \textbf{multi-class classifier}, each leaf node can be treated as a class. This classifier may divide the data space like the rectangle in Fig. \ref{DivideConquerFig} does and different colors represent different labels. Sometimes nearest neighbors (\eg white points in area $8$ and area $10$) are close to each other, but fit in different area according to this classifier with a discriminative plane (\eg node $2$) separating them. In Fig .\ref{DivideConquerFig}, suppose point $q$ will be assigned label $8$ when $q$ is input into the tree classifier. When conquering at level $1$, we need to know in subtree $5$ which area between node $10$ and $11$ is closer to $q$. Now that the whole tree can be treated as a \textbf{multi-class classifier}, any subtree of it can be a \textbf{multi-class classifier}, too. To know which area is closer, we simply let the classifier of subtree $5$ make the choice. By inputing $q$ to the classifier, we will obtain a label between 10 and 11 for q. Since $q$ is the white point in area $8$, from the rectangle in Fig. \ref{DivideConquerFig}, it's obvious that $q$ will be labeled as $10$. Therefore for $q$, when conquering at level $1$, only points in node $10$ will be involved. 

\begin{algorithm}
	\caption{Approximate $k$NN Graph Refinement Algorithm}
	\label{alg:generalRefine}
	\begin{algorithmic}[1]
		\Require an initial approximate $k$-nearest neighbor graph $G_{init}$, data set $D$, maximum iteration number $I$, Candidate pool size $P$, new neighbor checking num $L$.
		\Ensure an approximate $k$NN graph $G$.
		\State $iter = 0$, $G = G_{init}$
		\State Graph $G_{new}$ records all the new added candidate neighbors of each point. $G_{new} = G_{init}$. \State Graph $G_{old}$ records all the old candidate neighbors of each point at previous iterations. $G_{old} = \emptyset$
		\State Graph $G_{rnew}$ records all the new added reverse candidate neighbors of each point. \State Graph $G_{rold}$ records all the old reverse candidate neighbors of each point. 
		\While{$iter < I_{max}$}
		\State $G_{rnew} = \emptyset$, $G_{rold} = \emptyset$.
		\ForAll{point $i$ in $D$}
		\State $NN_{new}$ is the neighbor set of point $i$ in $G_{new}$.\State  $NN_{old}$ is the neighbor set of of point $i$ in $G_{old}$.
		\ForAll{point $j$ in $NN_{new}$}
		\ForAll{point $k$ in $NN_{new}$}
		\If{$j!=k$}
		\State calculate the distance between $j$ and $k$. \State add $k$ to $j$'s entry in $G$. mark $k$ as $new$. \State add $j$ to $k$'s entry in $G$ and $G_{rnew}$. \State mark $j$ as $new$.
		\EndIf
		\EndFor
		\ForAll{point $l$ in $NN_{old}$}
		\State calculate the distance between $j$ and $l$. \State add $l$ to $j$'s entry in $G$. mark $l$ as $old$.\State add $j$ to $l$'s entry in $G$ and $G_{rold}$.\State mark $j$ as $old$.
		\EndFor
		\EndFor
		\EndFor
		\ForAll{point $i$ in $D$}
		
		\State Reserve the closest $P$ points to $i$ in respective \State entry of $G$.
		
		\EndFor
		\State  $G_{new}=G_{old}=\emptyset$
		\ForAll{point $i$ in $D$}
		\State $l=0$.    $NN$ is the neighbor set of $i$ in $G$.
		\While{$l < L$ and $l < P$}
		\State $j = NN[ l ]$.
		\If{$j$ is marked as $new$}
		\State add $j$ to $i$'s entry in $G_{new}$.  \State $l = l + 1$.
		\Else
		\State add $j$ to $i$'s entry in $G_{old}$.
		\EndIf
		\EndWhile
		\EndFor
		\State $G_{new} = G_{new} \cup G_{rnew}$.\State $G_{old} = G_{old} \cup G_{rold}$
		
		\State $iter = iter + 1$
		\EndWhile
	\end{algorithmic}

\end{algorithm}

In this way, for each point at each level, only the points in one closest leaf node will be considered, which reduces the computation complexity greatly and reserves accuracy. We perform our random divide-and-conquer process multiple times and get an initial graph. 

Again there is a trade-off between accuracy of initial graph and time cost in parameter tuning. When conquer-to depth $Dep$ is small (\ie conquering to a level close to root), or when tree number $T_c$ is larger, the accuracy is higher but time cost is higher. In our experiments, we use the randomized KD-tree as the hierarchical divide-and-conquer structure. See Algorithm \ref{alg:generalBuild} for details on randomized KD-tree divide-and-conquer algorithm).

\subsubsection{Graph Refinement}

We use the NN-descent proposed by \cite{Dong2011Efficient} to refine the resulting graph we get from the divide-and-conquer step. The main idea is also to find better neighbors iteratively, however, different from NN-expansion, they proposed several techniques to get much better performance. We rewrite their algorithms to make it easy to understand. See Algorithm \ref{alg:generalRefine} for details. 

The pool size $P$ and neighbor checking num $L$ are essential parameters of this algorithm. Usually, Larger $L$ and $P$ will result in better accuracy but higher computation cost. 

\subsubsection{NN-expansion VS. NN-descent} \label{sec:NNexpNNdes}

Some approximate $k$NN graph construction methods \cite{Zhang2013Fast}  \cite{tang2016visualizing} claim to outperform NN-descent \cite{Dong2011Efficient} significantly. However, based on their reported results and our analysis, there seems a misunderstanding of NN-descent \cite{Dong2011Efficient}. Actually, NN-descent is quite different than NN-expansion. For given point $p$, NN-expansion assume the neighbors of $p$'s neighbors are likely to be neighbors of $p$. While NN-descent thinks that $p$'s neighbors are more likely to be neighbors of each other. Our experimental results have shown that NN-descent is much more efficient than NN-expansion in building approximate $k$NN graph. However, the NN-descent idea cannot be applied to ANN search.

\subsection{Online index updating}

EFANNA index building algorithm is easily to be extended to accept stream data. Firstly, when a new point arrived, we can insert it into the tree easily. And when the number of points in the inserted node exceeds given threshold, we just need to split the node. When the tree is unbalanced to some degree, we should adjust the tree structure, which is quite fast on large scale data. Secondly, the graph building algorithm can accept stream data as well, we can use the same algorithm we describe before. First we search in the tree for candidates, and use NN-descent to update the graph with the involved points. And this step is quite fast, too.

\section{Experiments}\label{sec4}

To demonstrate the effectiveness of the proposed method EFANNA, extensive experiments on  large-scale data sets are reported in this section. 

\subsection{Data Set and Experiment Setting}

The experiments were conducted on two popular real world data sets, SIFT1M and GIST1M\footnote{Both two datasets can be downloaded at http://corpus-texmex.irisa.fr/}. The detailed information on the data sets is listed in TABLE \ref{data_detail_table}. All the codes we used are written in C++ and compiled by g++4.9, and the only optimization option we allow is ``O3'' of g++. Parallelism and other optimization like SSE instruction are disabled. The experiment on SIFT1M is carried out on a machine with i7-3770K CPU and 16G memory, and GIST1M is on a machine with i7-4790K CPU and 32G memory.

\begin{table}[t]
\caption{information on experiment data sets}
\label{data_detail_table}
\centering
\begin{tabular}{|c|c|c|c|}
\hline
data set &  dimension & base number & query number \\
\hline
SIFT1M &  128 & 1,000,000 & 10,000 \\
GIST1M &  960 & 1,000,000 & 1,000 \\
\hline
\end{tabular}
\end{table}

\subsection{Experiments on ANN Search}
\label{simple_search_compare}
\subsubsection{Evaluation Protocol}

To measure the performance of ANN search of different algorithms, we used the well known 
$average$ $recall$ as the accuracy measurement \cite{Makhoul2000Performance}. Given a query point, all the algorithms are expected to return $k$ points. Then we need to examine how many points in this returned set are among the true $k$ nearest neighbors of the query. Suppose the returned set of $k$ points given a query is $R'$ and the true $k$ nearest neighbors set of the query is $R$, the $recall$ is defined as
\begin{equation}
recall(R') = \frac{|R' \cap R|}{|R|}.
\end{equation}
Then the $average$ $recall$ is averaging over all the queries. 
Since the sizes of $R'$ and $R$ are the same, the recall of $R'$ is the same as the accuracy of $R'$.

We compare the performance of different algorithms by requiring different number of nearest neighbors of each query point, including 1-NN and 100-NN. In other words, the size of $R$ (and $R'$) will be 1 and 100 respectively. Please see our technical report \cite{CongEfanna2016} for more results on 10-NN and 50-NN.

\begin{figure*}
	\centering
	\subfigure[SIFT1M 1NN]{\includegraphics[width=\DoubleFigureWidth]{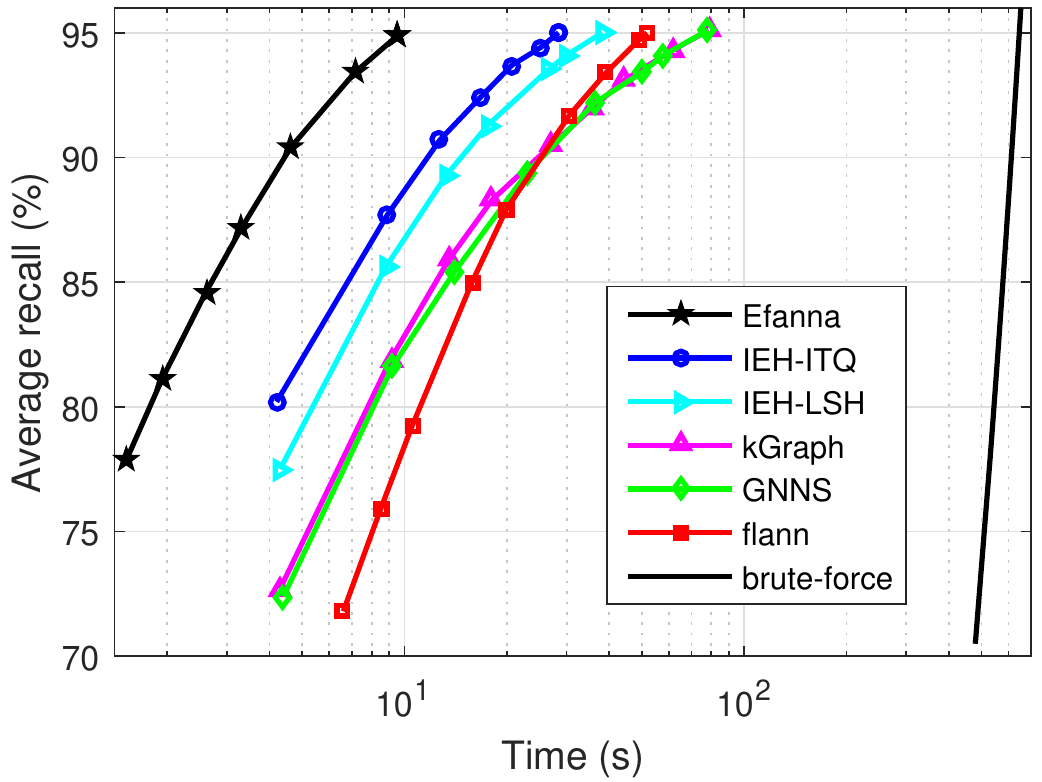}}
	\subfigure[SIFT1M 10NN]{\includegraphics[width=\DoubleFigureWidth]{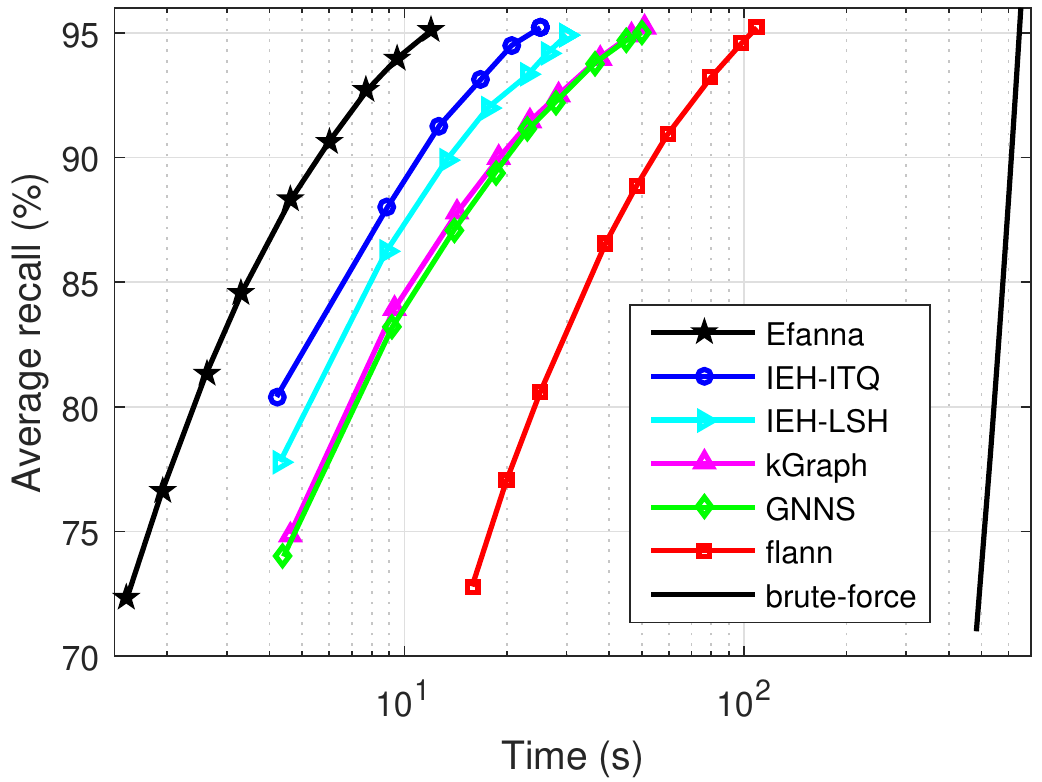}}
	\subfigure[SIFT1M 50NN]{\includegraphics[width=\DoubleFigureWidth]{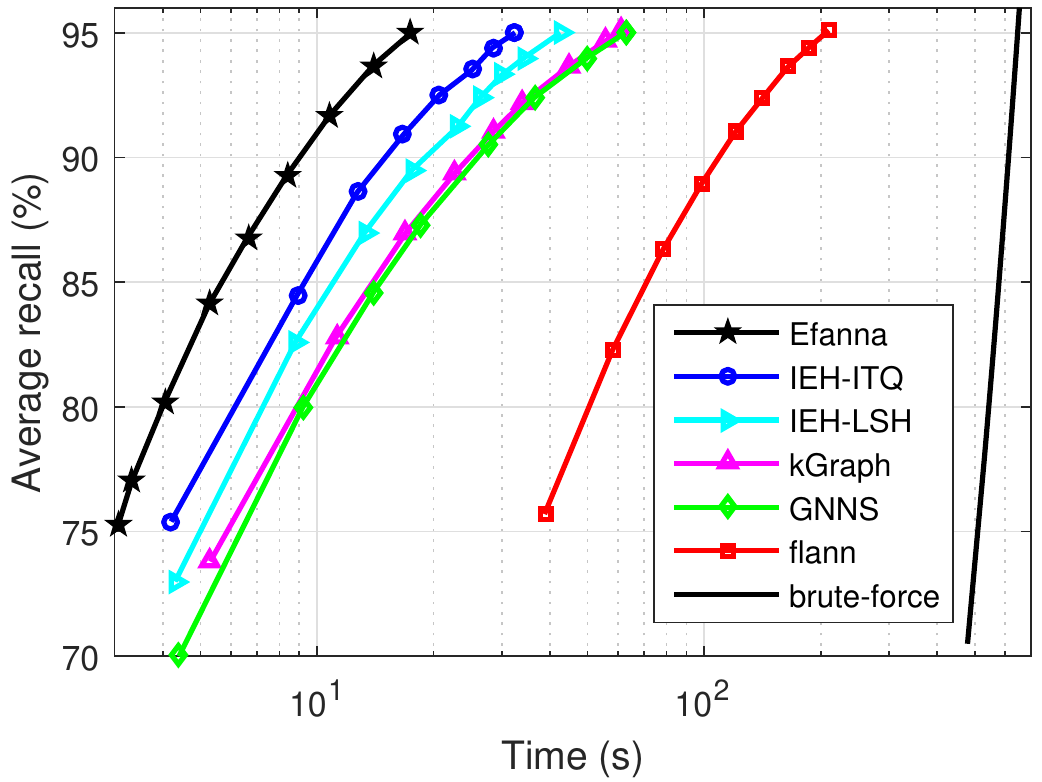}}
	\subfigure[SIFT1M 100NN]{\includegraphics[width=\DoubleFigureWidth]{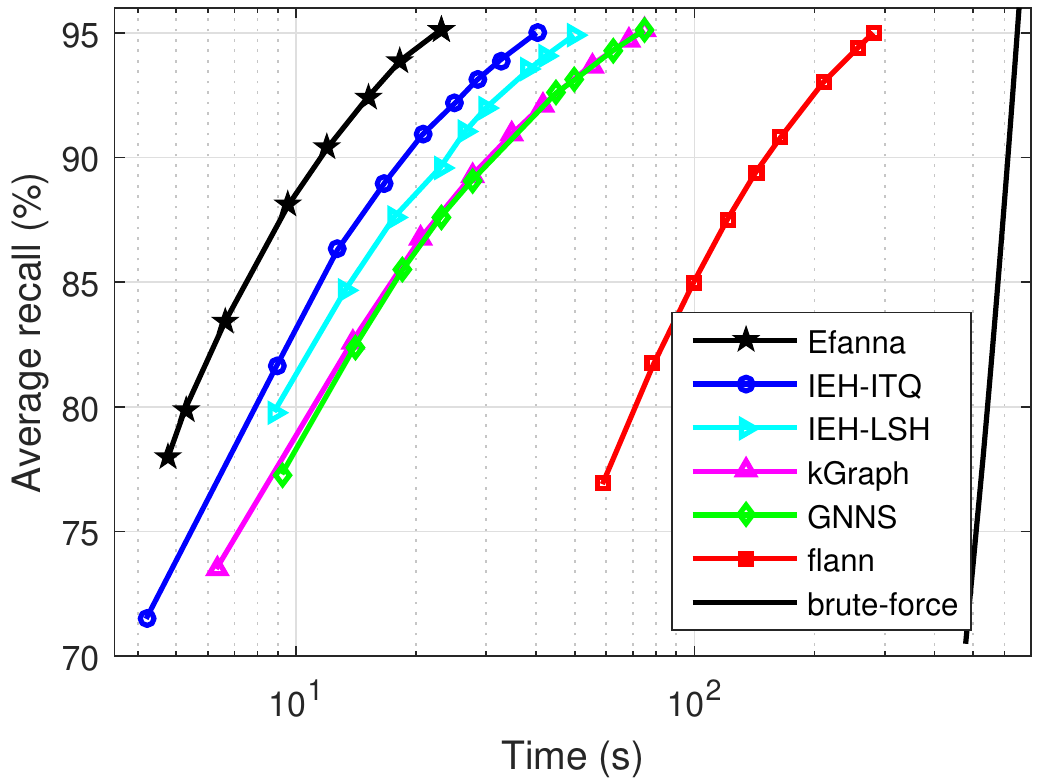}}
	\caption{ANN search results of 10,000 queries on SIFT1M. We use a 10-NN ground truth graph for all graph based methods.}
	\label{SIFT_search_gt}
\end{figure*}

\begin{figure*}
	\centering
	\subfigure[GIST1M 1NN]{\includegraphics[width=\DoubleFigureWidth]{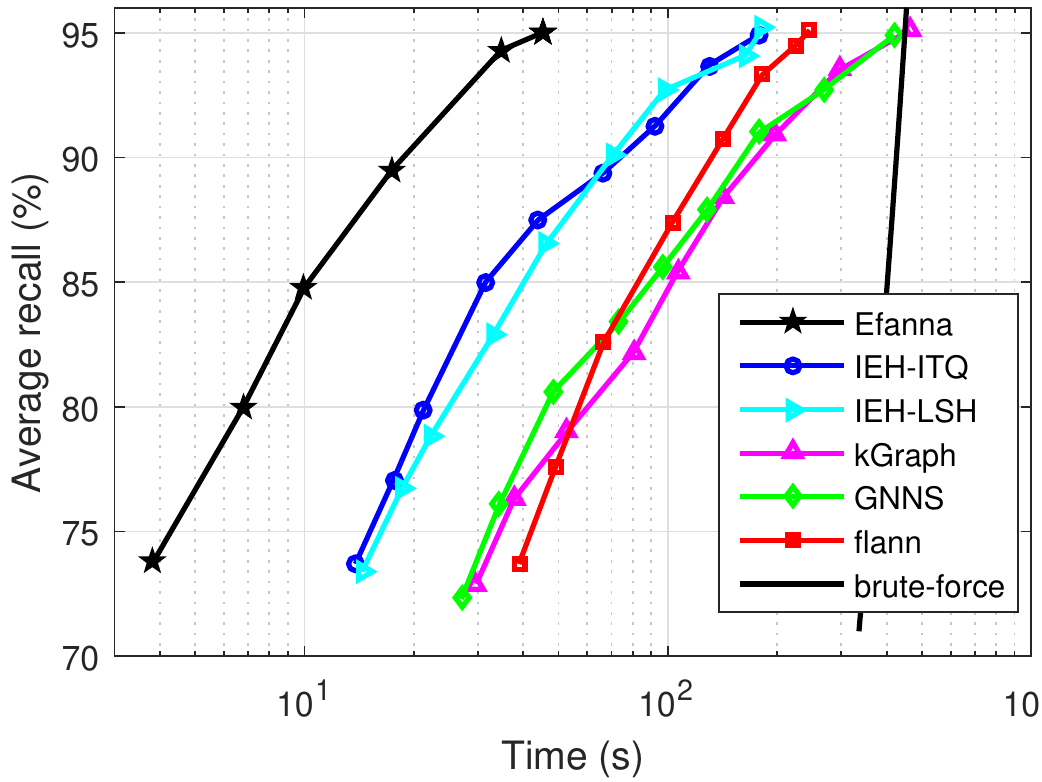}}
	\subfigure[GIST1M 10NN]{\includegraphics[width=\DoubleFigureWidth]{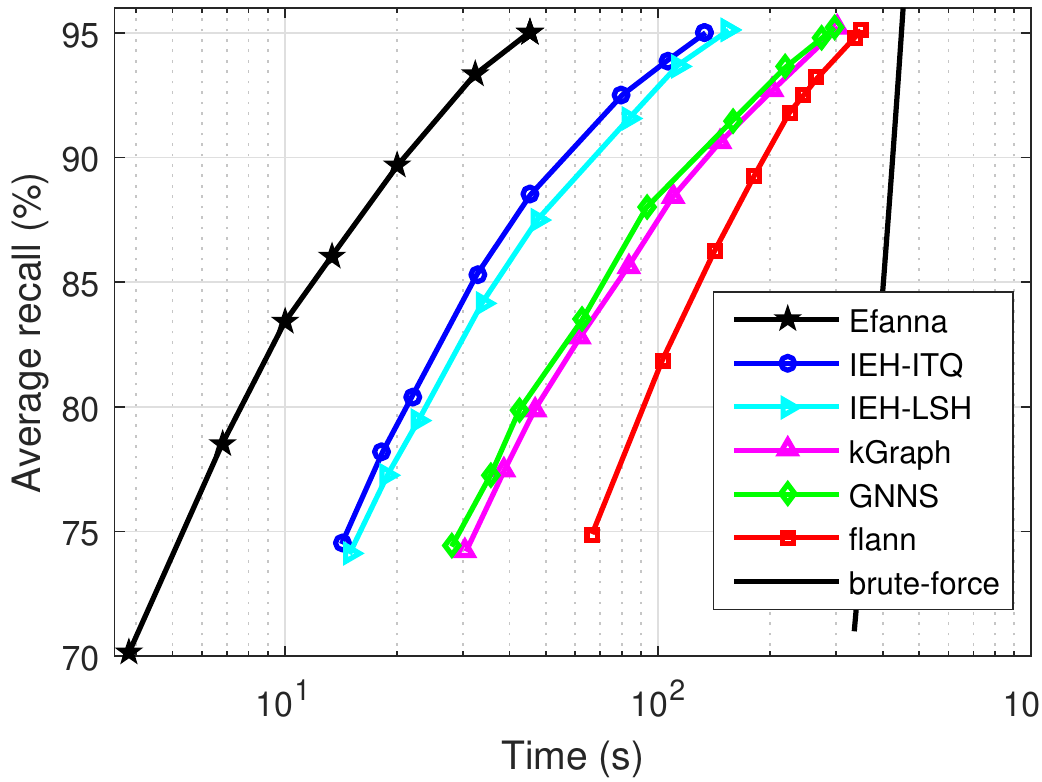}}
	\subfigure[GIST1M 50NN]{\includegraphics[width=\DoubleFigureWidth]{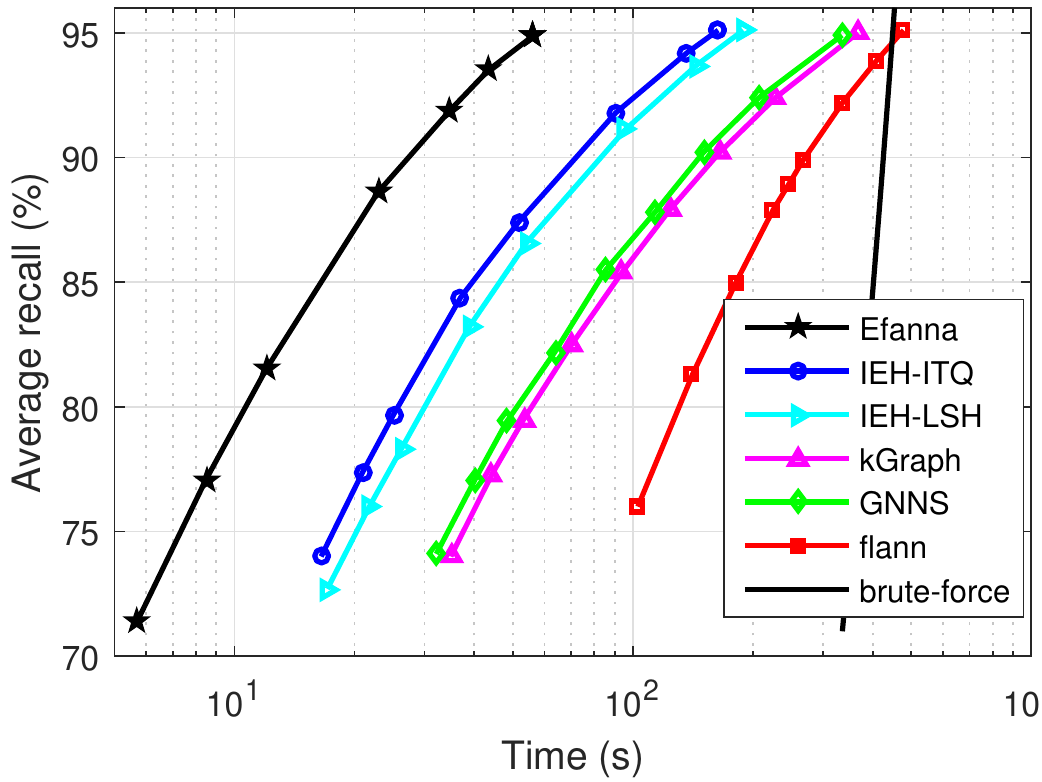}}
	\subfigure[GIST1M 100NN]{\includegraphics[width=\DoubleFigureWidth]{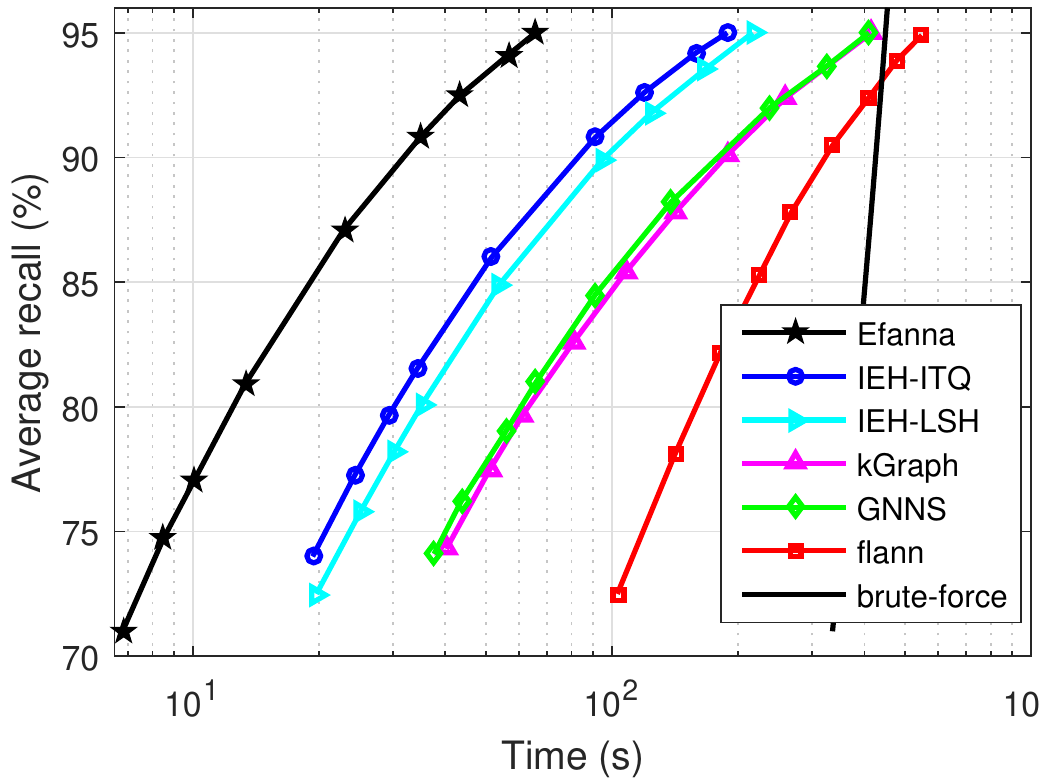}}
	\caption{ANN search results of 1,000 queries on GIST1M. We use a 10-NN ground truth graph for all graph based methods.}
	\label{GIST_search_gt}
\end{figure*}

\subsubsection{Comparison Algorithms}

To demonstrate the effectiveness of the proposed EFANNA approach, the following four state-of-the-art ANN search methods and brute-force method are compared in the experiment.

\begin{enumerate}
\item \textbf{brute-force}. We report the performance of brute-force search to show the advantages of using ANN search methods. To get different recall, we simply perform  brute-force search on different percentage of the query number. For example, the brute-force search time on 90\% queries of the origin query set stands for the brute-force search time of 90\% average recall. 
	
\item \textbf{flann}. FLANN is a well-known open source library for ANN search \cite{Muja2014Scalable}. The Randomized KD-tree algorithm in FLANN provides state-of-the-art performance. In our experiments, we use 16 trees for both datasets. And we tune the ``max-check'' parameter to get the time-recall curve.

\item \textbf{GNNS}. GNNS is the first ANN search method using $k$NN graph \cite{Hajebi2011Fast}.  Given a query, GNNS generates the initial candidates (neighbors) by random selection. Then GNNS uses the NN-expansion idea (\ie, check the neighbors of the neighbors iteratively to locate closer neighbors) to refine the result. The main parameters of GNNS are the size of the initial result and the iteration number. We fix the iteration number as 10 and tune the initial candidate number to get the time-recall curve.

\item \textbf{kGraph}. kGraph \cite{Kgraph2014} is an open library for ANN search based on $k$NN graph. The author of kGraph is the inventor of NN-descent \cite{Dong2011Efficient}. The ANN search algorithm in kGraph is essentially the same as GNNS. The original Kgraph library implements with OpenMP (for parallelism) and SSE instructions for speed-up. We simply turn off the parallelism and SSE for fair comparison.

\item \textbf{IEH}. IEH is a short name for Iterative Expanding Hashing \cite{Jin2014Fast}. It is another ANN search method using $k$NN graph. Different from GNNS, IEH uses hashing methods to generate the initial result given a query. Considering the efficiency of hash coding,  \textbf{IEH-LSH} and \textbf{IEH-ITQ} are compared in our experiment. The former uses LSH \cite{Gionis1999Similarity} as the hashing method and the latter uses ITQ \cite{Yunchao2011Iterative} as the hashing method. Both hashing methods use 32 bit code. We also fix the iteration number as 10 and tune the initial result size to get the time-recall curve.

\item \textbf{Efanna}. The algorithm proposed in this paper. We use 16 trees for both datasets and the iteration number in NN-expansion stage is fixed as 4. We tune the search-to depth parameter $S_{depth}$ and the candidate pool size $P$ to get the time-recall curve.
\end{enumerate}

\begin{table*}[t]
	\caption{Index Size of Different Algorithms}
	\label{IndexSizeTable}
	\centering
	\begin{tabular}{|c|c|c|c|c|}
		\hline
		\multirow{2}{*}{data set} & \multirow{2}{*}{algorithms}& \multicolumn{3}{c|}{ index size} \\
		\cline{3-5}
		& & tree (hash table) & graph & all \\
		\hline 
		\multirow{6}{1cm}{SIFT1M}  
		&flann(16-tree)& 997.5 MB & 0 & 997.5 MB\\
		\cline{2-5}
		&Efanna(16-tree, 10-NN)& 283.3 MB & 60.5 MB & 343.8 MB\\
		\cline{2-5}
		&GNNS(10-NN)& 0 & 60.5 MB & 60.5 MB \\
		\cline{2-5}
		&kGraph (10-NN)&  0 & 60.5 MB & 60.5 MB \\
		\cline{2-5}
		&IEH-LSH (32 bit, 10-NN)&  82.7 MB & 60.5 MB & 143.2 MB \\
		\cline{2-5}
		&IEH-ITQ (32 bit, 10-NN)& 82.7 MB & 60.5 MB & 143.2 MB \\
		\hline
		\multirow{6}{1cm}{GIST1M}  
		&flann(16-tree)& 998.4 MB & 0 & 998.4 MB\\
		\cline{2-5}
		&Efanna(16-tree, 10-NN)& 287.7 MB & 60.5 MB & 348.2 MB\\
		\cline{2-5}
		&GNNS(10-NN)& 0 & 60.5 MB & 60.5 MB \\
		\cline{2-5}
		&kGraph (10-NN)&  0 & 60.5 MB & 60.5 MB \\
		\cline{2-5}
		&IEH-LSH (32 bit, 10-NN)&  82.7 MB & 60.5 MB & 143.2 MB \\
		\cline{2-5}
		&IEH-ITQ (32 bit, 10-NN)& 82.7 MB & 60.5 MB & 143.2 MB \\
		\hline
		\multicolumn{5}{|l|}{The index size here is the size in the memory, not the size on the disk.}\\
		\hline
	\end{tabular}
\end{table*}

Among the five compared ANN methods, Flann's KD-tree is the hierarchical structure (tree) based method. The other four compared methods are all graph based methods. We do not compare with hashing based methods because \cite{Jin2014Fast} shows the significant improvement of IEH over the corresponding hashing methods.

All the graph based methods need a pre-built $k$NN graph and we use a ground truth 10-NN graph.

\subsubsection{Results}

The time-recall curves of all the algorithms on two data sets can be seen in Fig. \ref{SIFT_search_gt} and Fig. \ref{GIST_search_gt}. The index size of various algorithms are shown in Table \ref{IndexSizeTable}. A number of interesting conclusions can be drawn as follows:

\begin{enumerate}
	\item Our Efanna algorithm significantly outperforms all the other methods at all the cases on both of two data sets. Even at a relatively high recall (\eg, 95\%), Efanna is about 100x faster than the brute-force search on the SIFT1M and about 10x faster than the brute-force search on the GIST1M.  
	
	\item The GIST1M is a harder dataset than the SIFT1M for ANN search. At a relatively high recall (\eg, 95\%), all the ANN search methods are significantly faster than the brute-force search. However, on GIST1M some methods (flann, GNNS, kGrpah) are similar (or even slower) to the brute-force search. The reason may be the high dimensionality of the GIST1M.
	
	\item When the required number of nearest neighbors is large (\eg, 10, 50 and 100), all the graph based methods are significantly better than Flann's KD-tree. Since 10, 50 or 100 results are more common in practical search scenarios, the graph based methods have the advantage. 
	
	\item GNNS and kGraph are essentially the same algorithm. The experimental results confirm this. The slight difference may due to the random initialization.  
	
	\item We implement four graph based methods (GNNS, IEH-LSH, IEH-ITQ and Efanna) exactly with the same framework. The only difference is the initialization: GNNS uses the random selection, IEH uses the hashing and Efanna uses the truncated KD-tree. The performance gap between these methods indicates the effectiveness of different initialization methods. The truncated KD-tree is better than the hashing and these two are better than the random selection. And the ITQ is better than the LSH.
	
	\item TABLE \ref{IndexSizeTable} shows the index size of different algorithms. Flann consumes the largest memory size. The index size of Efanna is slightly larger than IEH. To reduce the index size, one can use less trees in Efanna but maintain the high performance. We will discuss this in the section \ref{TreeSection}.
	
	\item The index size of GNNS and KGraph is smallest because they only need to store a $k$NN graph. Both IEH and Efanna sacrifice the index size (additional data structure for better initialization) for better search performance. 
	
	\item Considering both search performance and index size, graph based methods is a better choice than Flann's KD-tree.
\end{enumerate}

\subsection{Experiment on Approximate kNN Graph Construction}

We show in last section that graph based methods can achieve very good performance on ANN search. However, the results above are based on a ground truth 10-NN graph. Table \ref{bfsearch_table} shows the time cost to build the ground truth 10-NN graph for two datasets. It takes about 17 hours of CPU time on SIFT1M and about a week on GIST1M.  Obviously, brute-force is not an acceptable choice. \cite{Hajebi2011Fast,Jin2014Fast} assume that the ground truth $k$NN graph exists. However, building the $k$NN graph is a step of indexing part of all the graph based methods. To make the graph based ANN search methods practically useful, we need to discuss how to build the $k$NN graph efficiently.

In this section, we will compare the performance of several approximate $k$NN graph construction methods.

\subsubsection{Evaluation Protocol}

We use the accuracy-time curve to measure the performance of different approximate $k$NN graph construction algorithms. Given a data set with $N$ points, an approximate $k$NN graph construction method should return $N$ groups of $k$ points, and each group of points stands for nearest neighbors the algorithm finds within the data set for the respective point. Suppose for point $i$, the returned point set of  is $R'_i$ and the ground truth set is $R_i$. Then the accuracy of point $i$, $accuracy_i$, is defined as 
\begin{equation}\label{eq:accuracy}
accuracy_i = \frac{|R'_i \cap R_i|}{|R'_i|}
\end{equation}
Then the $Accuracy$ of the returned graph is defined as the average accuracy over all the $N$ points:
\begin{equation}
Accuracy = \frac{\sum_{i}^{N}|R'_i \cap R_i|}{N|R'_i|}
\end{equation}

We compare the performance of all the algorithms on building a $10$-NN graph (\ie, the sizes of $R_i$ and $R'_i$ are 10).

\begin{table}[t]
	\caption{Time of building the ground truth 10-NN graph on GIST1M and SIFT1M using brute-force search}
	\label{bfsearch_table}
	\centering
	\begin{tabular}{|p{2.5cm}<{\centering}|p{3.5cm}<{\centering}|}
		\hline
		data set & time (seconds) \\
		\hline
		SIFT1M & 68,060 \\
		GIST1M & 565,060 \\
		\hline
	\end{tabular}
\end{table}

\begin{figure}[t]
	\centering
	\includegraphics[width=3.5in]{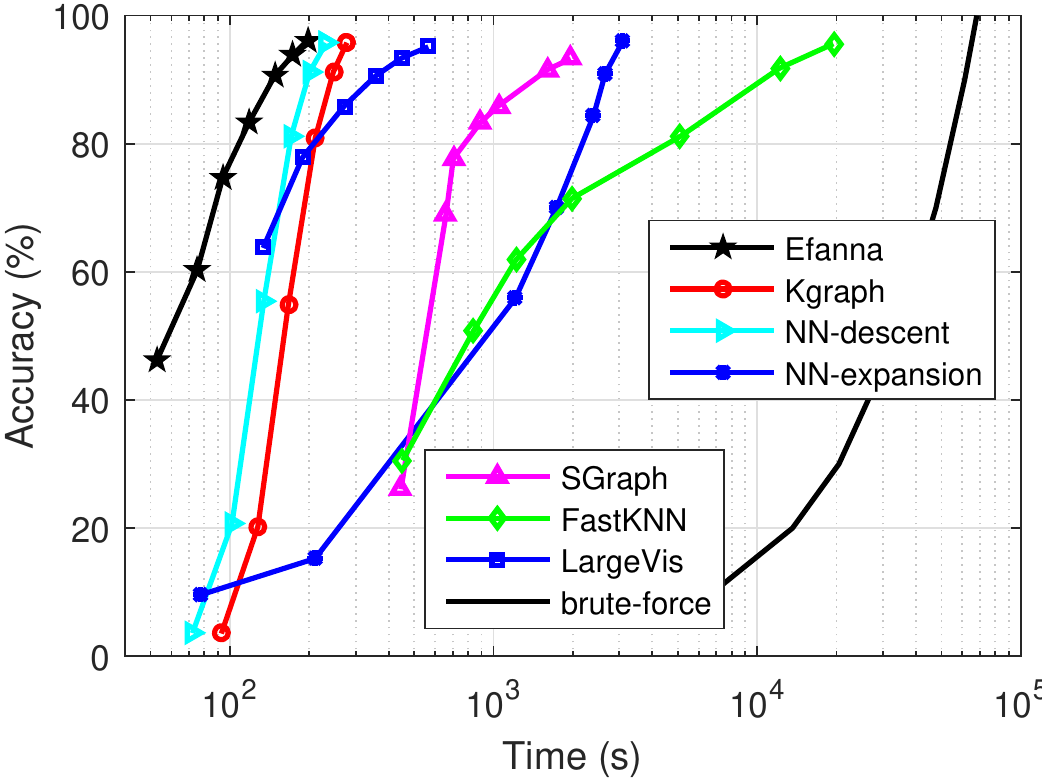}
	\caption{10-NN approximate graph construction results on SIFT1M}
	\label{SIFT_graph}
\end{figure}

\begin{figure}[t]
	\centering
	\includegraphics[width=3.5in]{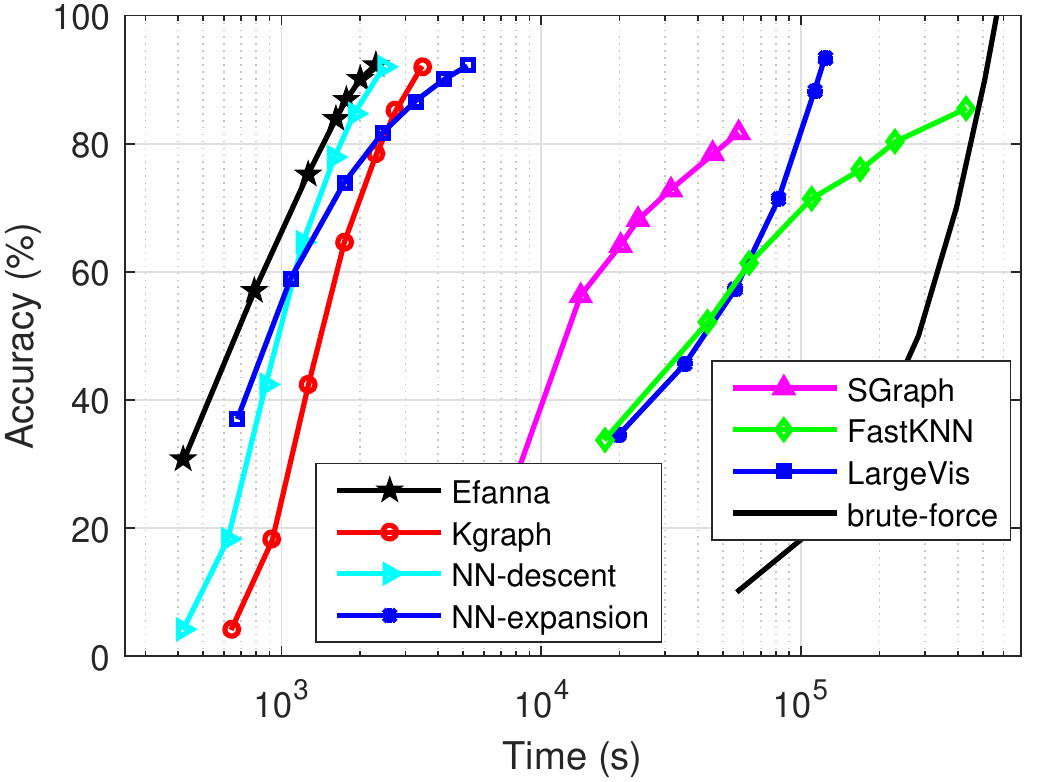}
	\caption{10-NN approximate graph construction results on GIST1M}
	\label{GIST_graph}
\end{figure}

\subsubsection{Comparison Algorithms}

\begin{enumerate}

\item \textbf{brute-force}: We report the performance of brute-force graph construction to show the advantages of using approxiamate $k$NN graph construction methods. To get different graph accuracy, we simply perform brute-force graph construction on different percentage of the data points. 

\item \textbf{SGraph}: We refer to the algorithm proposed in \cite{Gan2012Scalable} as SGraph. SGraph build the graph with three steps. First they generates initial graph by randomly dividing the data set into small ones iteratively and the dividing is carried out many times. Then they do brute-force graph construction within each subsets and combine all the subgraph into a whole. Finally they refine the graph using a technique similar to NN-expansion.

\item \textbf{FastKNN}: We refer to the algorithm proposed in \cite{Zhang2013Fast} as FastKNN. The last two steps of their graph building process is similar to SGraph. While FastKNN uses hashing method (specifically, AGH\cite{liu2011hashing}) to generate the initial graph.

\item \textbf{NN-expansion}: The main idea of building approximate $k$NN graph with NN-expansion is to cast the graph construction problem as $N$ ANN search problems, where $N$ is the data size. However, NN-expansion is proposed for ANN search while not for AKNN graph construction. The reason we add it to the compared algorithms in this section is that some previous works  \cite{Zhang2013Fast}  \cite{tang2016visualizing} claim to outperform NN-descent. While we find there may be misunderstanding that they may actually compared with NN-expansion rather than NN-descent. 

\item \textbf{NN-descent} \cite{Dong2011Efficient}: This algorithm first initializes the graph randomly. Then NN-descent refine it iteratively with techniques like local join and sampling \cite{Dong2011Efficient}. Local join is to do brute-force searching within a point $q$'s neighbors which is irrelevant to $q$. Sampling is to ensure number of points involved in the local join is small, but the algorithm is still efficient. 

\item \textbf{kGraph}: kGraph \cite{Kgraph2014} is an open source library for approximate $k$NN graph construction and ANN search. The author of kGraph is the author of NN-descent. The approximate $k$NN graph algorithm implemented in kGraph library is exactly NN-descent. kGraph implements with OpenMP and SSE instruction for speed-up. For faire comparison, we disable the parallelism and SSE instruction. 

\item \textbf{LargeVis} \cite{tang2016visualizing}: This algorithm is proposed for high dimension data visualization. The first step of LargeVis is to build an approximate $k$NN graph. LargeVis uses random projection tree and NN-expansion to build this graph. 

\item \textbf{Efanna}: The algorithm proposed in this paper. We use hierarchical divide-and-conquer to get an initial graph. And then use NN-descent to refine the graph. In this experiments, we use 8 randomized truncated KD-trees to initialize the graph.

\end{enumerate}

\begin{figure*}
	\centering
	\subfigure[SIFT1M 1NN]{\includegraphics[width=\DoubleFigureWidth]{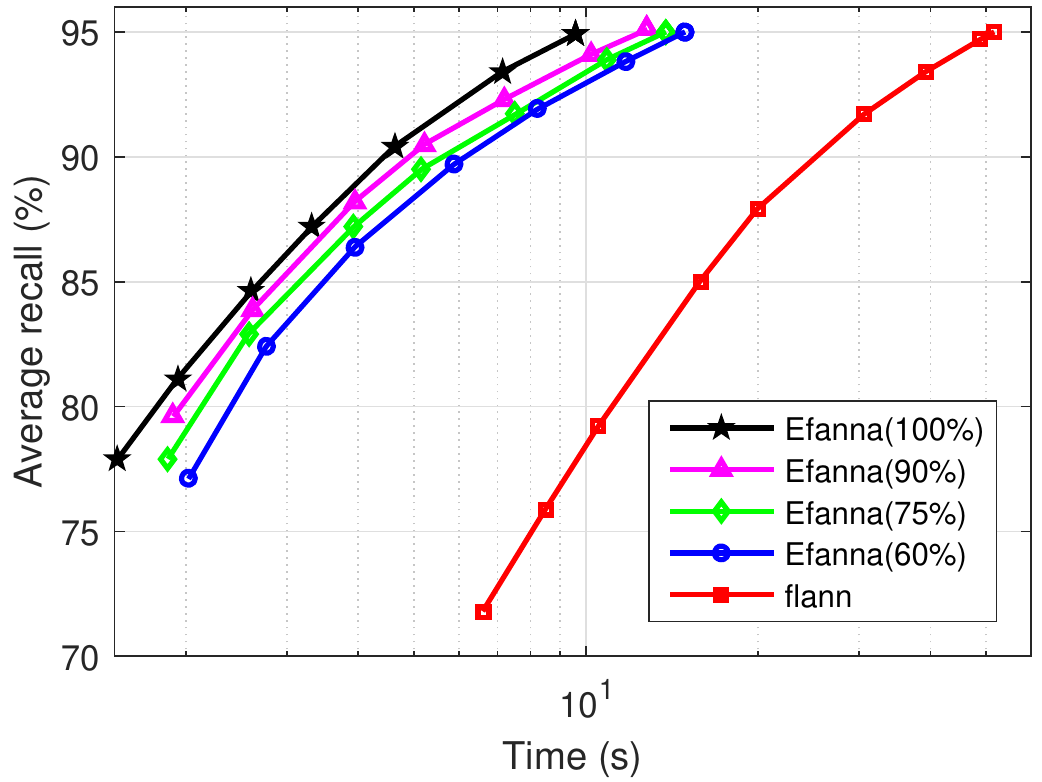}}
	\subfigure[SIFT1M 10NN]{\includegraphics[width=\DoubleFigureWidth]{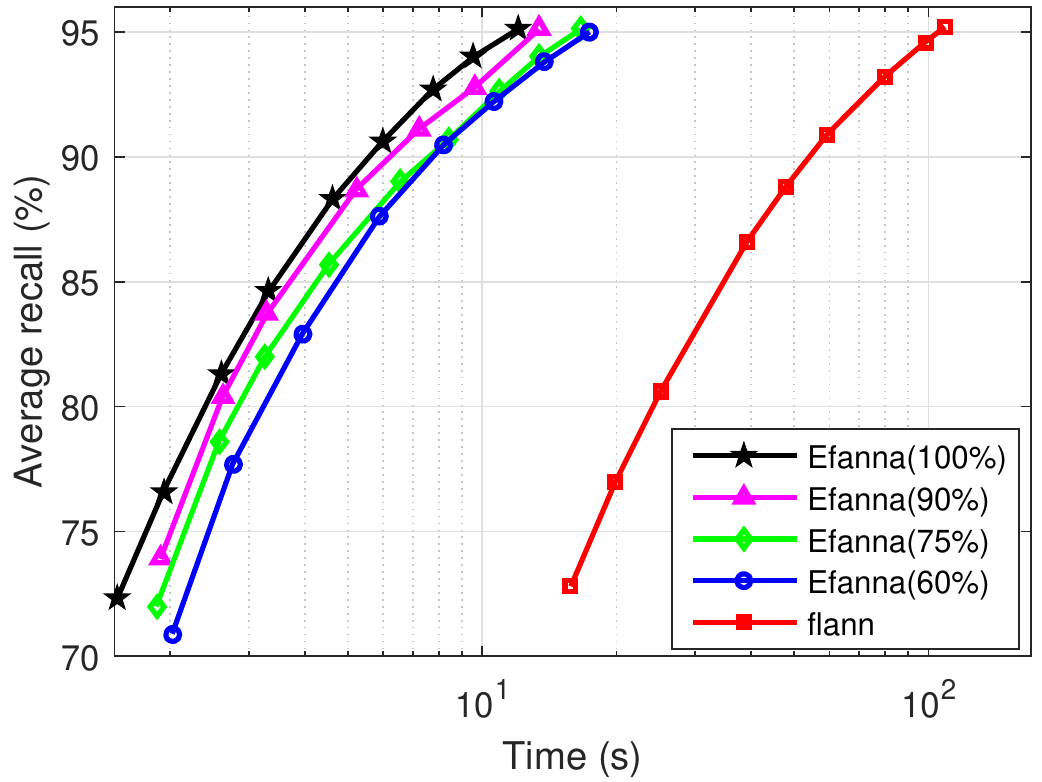}}
	\subfigure[SIFT1M 50NN]{\includegraphics[width=\DoubleFigureWidth]{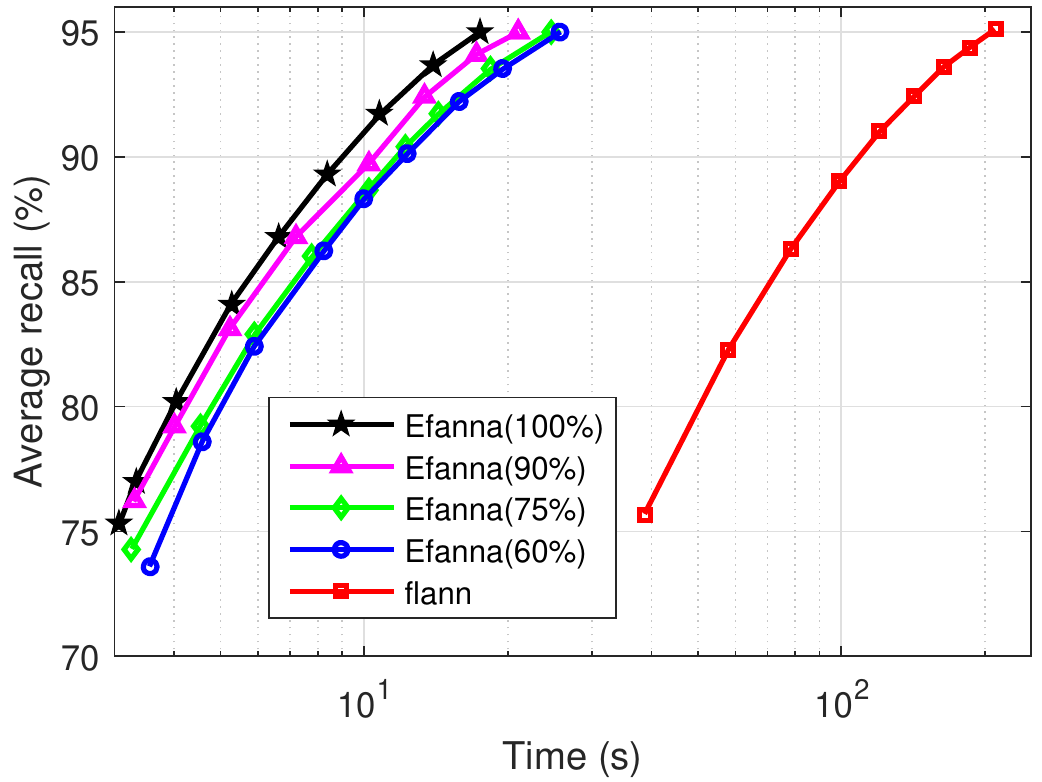}}
	\subfigure[SIFT1M 100NN]{\includegraphics[width=\DoubleFigureWidth]{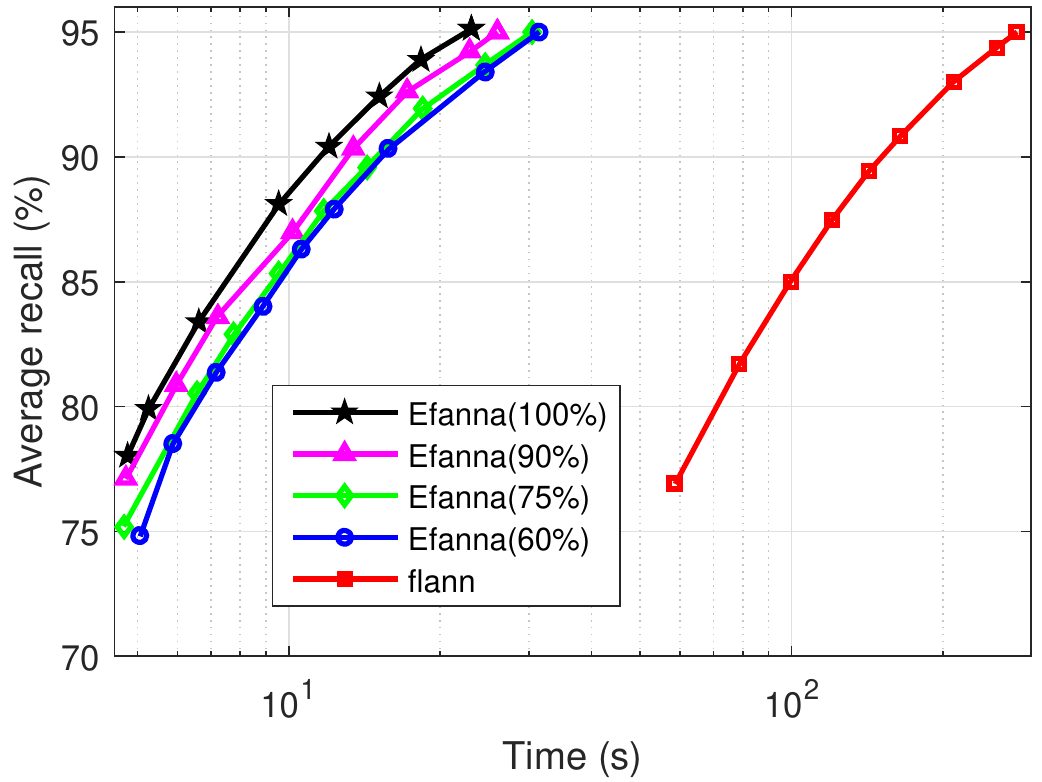}}
	\caption{Approximate nearest neighbor search results of 10,000 queries on SIFT1M. We use a 60\% $\sim$ 100\% accuracy 10-NN graphs for EFANNA respectively; Both EFANNA and flann-kdtrees use 16 trees.}
	\label{SIFT_search_graph}
\end{figure*}

\begin{figure*}
	\centering
	\subfigure[GIST1M 1NN]{\includegraphics[width=\DoubleFigureWidth]{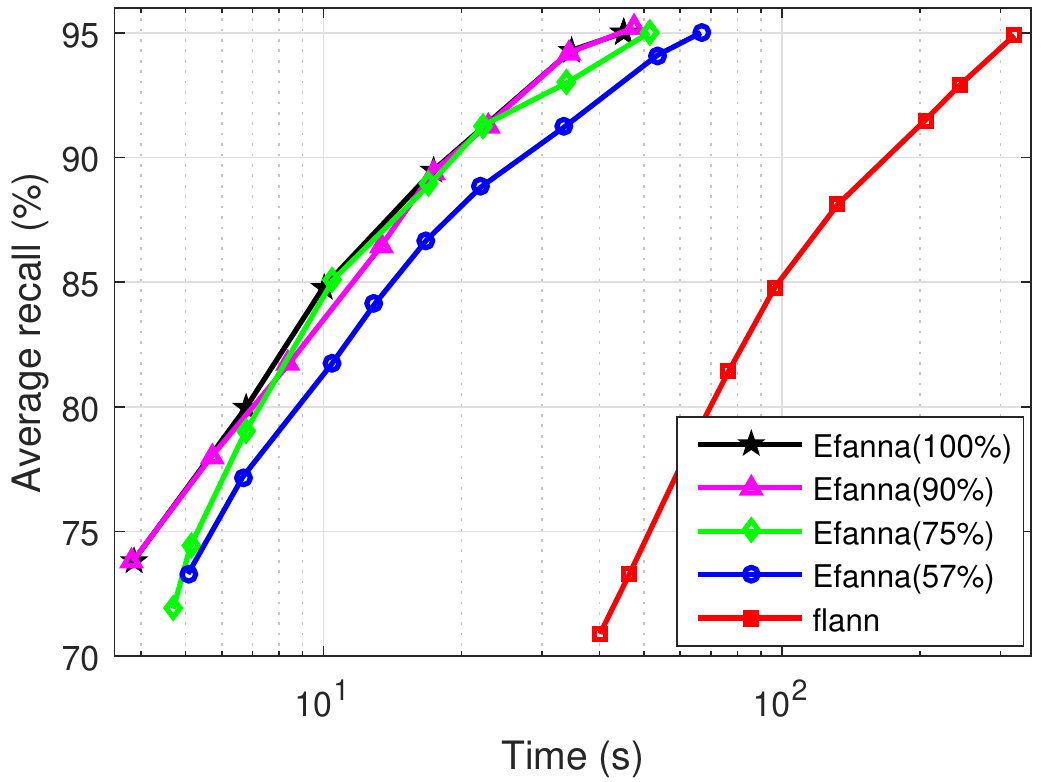}}
	\subfigure[GIST1M 10NN]{\includegraphics[width=\DoubleFigureWidth]{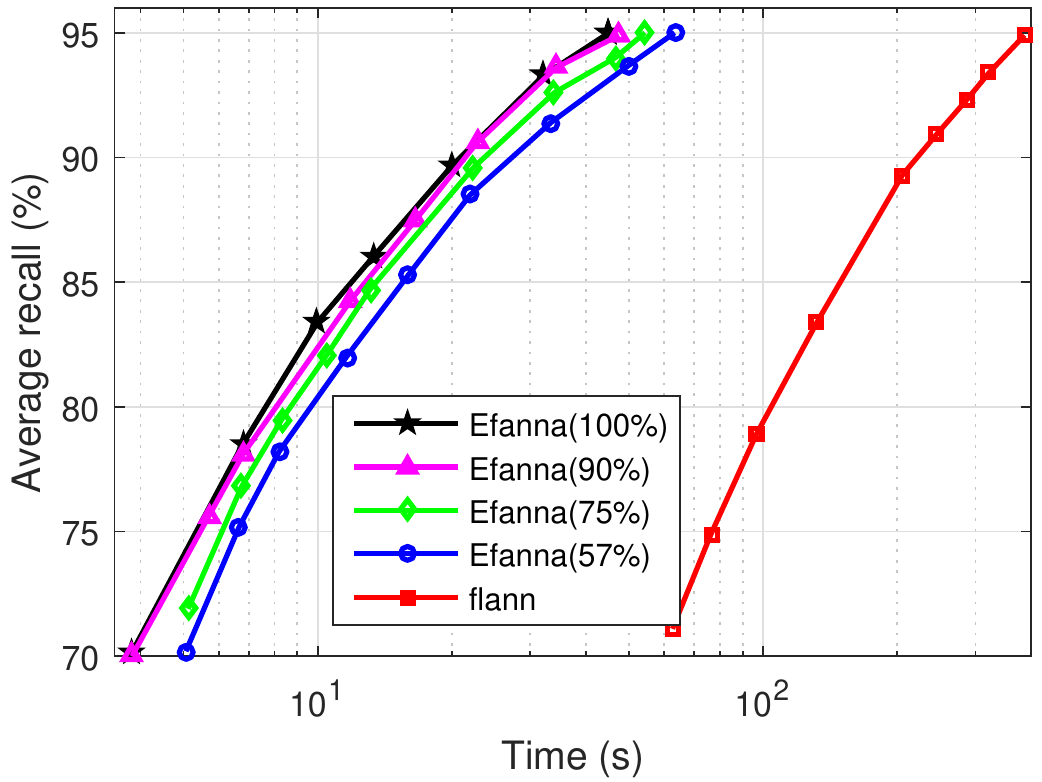}}
	\subfigure[GIST1M 50NN]{\includegraphics[width=\DoubleFigureWidth]{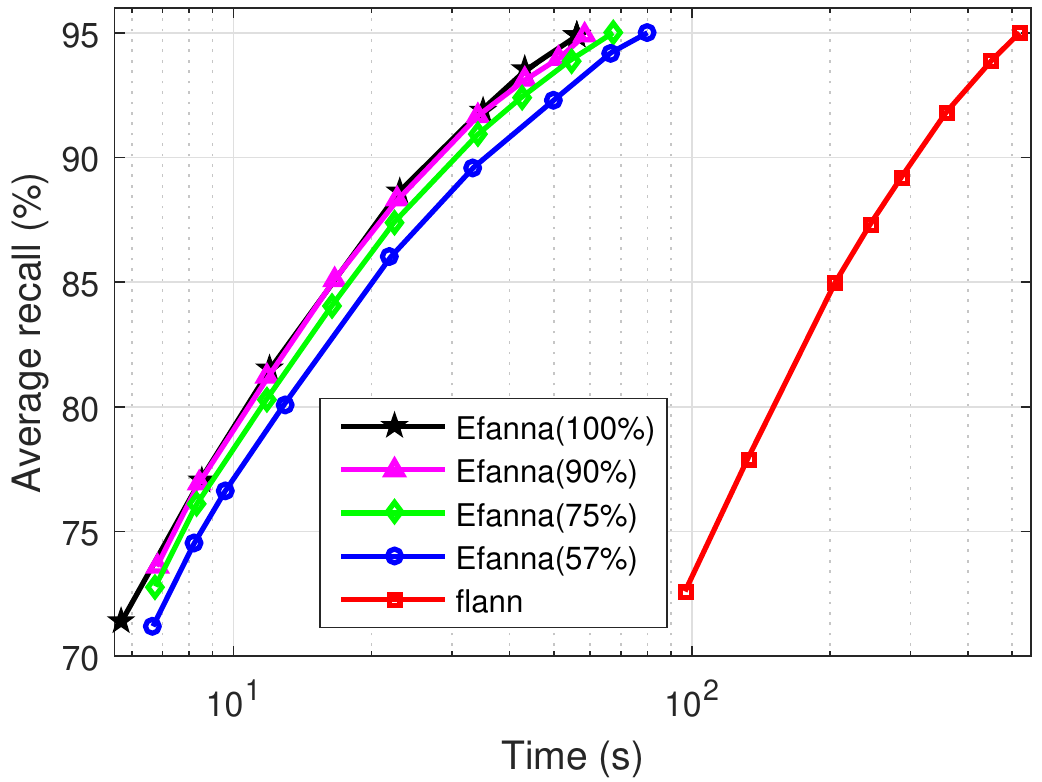}}
	\subfigure[GIST1M 100NN]{\includegraphics[width=\DoubleFigureWidth]{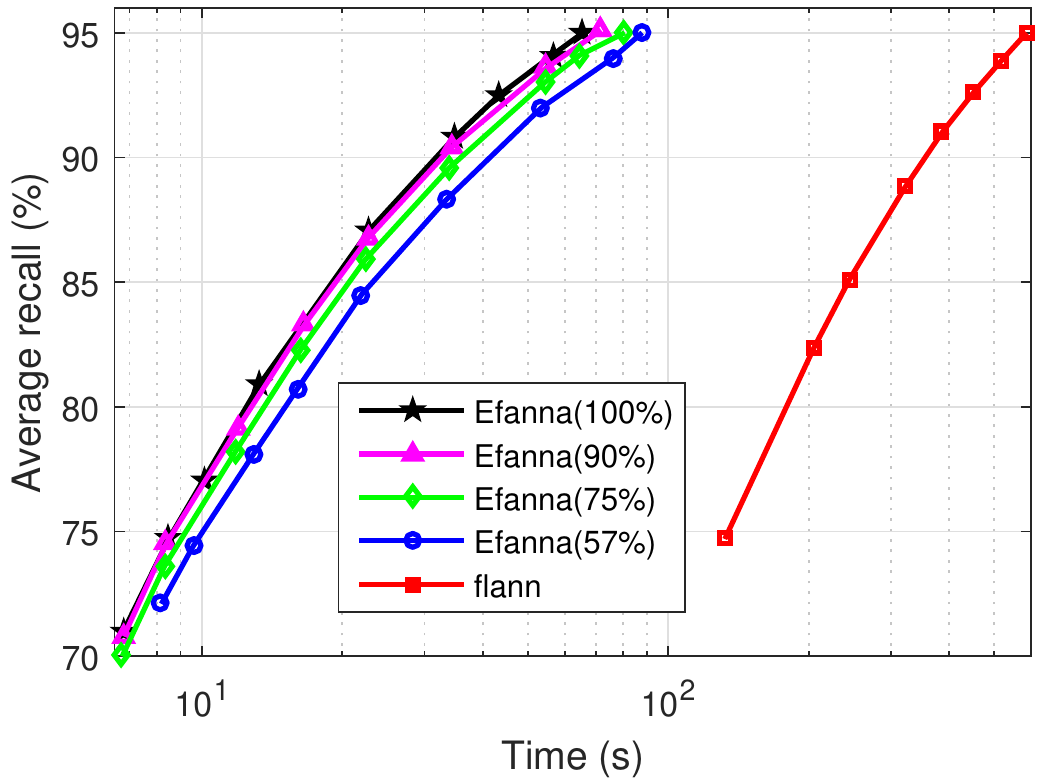}}
	\caption{Approximate nearest neighbor search results of 1,000 queries on GIST1M. We use a 57\% $\sim$ 100\% accuracy 10-NN graphs for EFANNA respectively; Both EFANNA and flann-kdtrees use 16 trees.}
	\label{GIST_search_graph}
\end{figure*}

\begin{table*}[t]
	\caption{Indexing Time of Efanna and flann}
	\label{IndexTimeTable}
	\centering
	\begin{tabular}{|c|c|c|c|c|}
		\hline
		\multirow{2}{*}{data set} & \multirow{2}{*}{algorithms}& \multicolumn{3}{c|}{ indexing time} \\
		\cline{3-5}
		& & tree building & graph building & all \\
		\hline 
		\multirow{4}{1cm}{SIFT1M}  
		&flann(16-tree)& 131.7s & 0 & 131.7s\\
		\cline{2-5}
		&Efanna(16-tree, 90\% 10-NN)& 8.2s & 148.0s & 156.2s\\
		\cline{2-5}
		&Efanna(16-tree, 75\% 10-NN)& 8.2s & 93.9s & 102.1s\\
		\cline{2-5}
		&Efanna(16-tree, 60\% 10-NN)& 8.2s & 75.1s & 83.3s\\
		\hline
		\multirow{4}{1cm}{GIST1M}  
		&flann(16-tree)& 707.5s & 0 & 707.5s\\
		\cline{2-5}
		&Efanna(16-tree, 90\% 10-NN)& 22.6s & 2017.7s & 2040.3s\\
		\cline{2-5}
		&Efanna(16-tree, 75\% 10-NN)& 22.6s & 1267.7s & 1290.3s\\
		\cline{2-5}
		&Efanna(16-tree, 57\% 10-NN)& 22.6s & 788.7s & 811.3s\\
		\hline
	\end{tabular}
\end{table*}

\begin{table*}[t]
	\caption{Efanna graph accuracy VS. k}
	\label{GraphAccuracyTable}
	\centering
	\begin{tabular}{|c|c|c|c|c|c|c|c|c|c|c|c|}
		\hline
		\multirow{2}{*}{data set} & \multirow{2}{*}{Efanna graph}& \multicolumn{10}{c|}{Accuracy} \\
		\cline{3-12}
		& & 10NN & 20NN & 30NN & 40NN & 50NN & 60NN& 70NN & 80NN & 90NN & 100NN\\
		\hline 
		\multirow{3}{1cm}{SIFT1M}  
		& 90\% 10-NN& 0.906515 &  0.99165 &  0.99788 &  0.999225 & 0.999654 & 0.999826 &0.999901 &0.999936 &0.999955 &0.999968\\
		\cline{2-12}
		&75\% 10-NN& 0.747126 &  0.932613 & 0.971548 & 0.985388 & 0.991626 & 0.994829 & 0.996636 &0.997721 & 0.998397 &  0.998838 \\
		\cline{2-12}
		&60\% 10-NN&  0.602044& 0.823565  & 0.899977 & 0.936402 &0.956686  & 0.969043 &0.977085  &  0.982535 &0.986419 &  0.989237 \\
		\hline
		\multirow{3}{1cm}{GIST1M}  
		& 90\% 10-NN&0.901412  & 0.994701 & 0.998917  &  0.999624 & 0.999828 & 0.999896 &0.999921  & 0.999932 & 0.99994 & 0.999943 \\
		\cline{2-12}
		&75\% 10-NN& 0.75174  & 0.937751  &  0.974985 &0.987682  &0.993141  & 0.99586  & 0.99735 & 0.998223 &0.998756  & 0.999094 \\
		\cline{2-12}
		&57\% 10-NN& 0.570855  & 0.792676  & 0.876249  & 0.91897  & 0.943883  &  0.959537 &  0.969966 &0.97717  &0.982319  & 0.986129\\
		\hline
	\end{tabular}
\end{table*}

\begin{figure*}
	\centering
	\subfigure[SIFT1M 1NN]{\includegraphics[width=\DoubleFigureWidth]{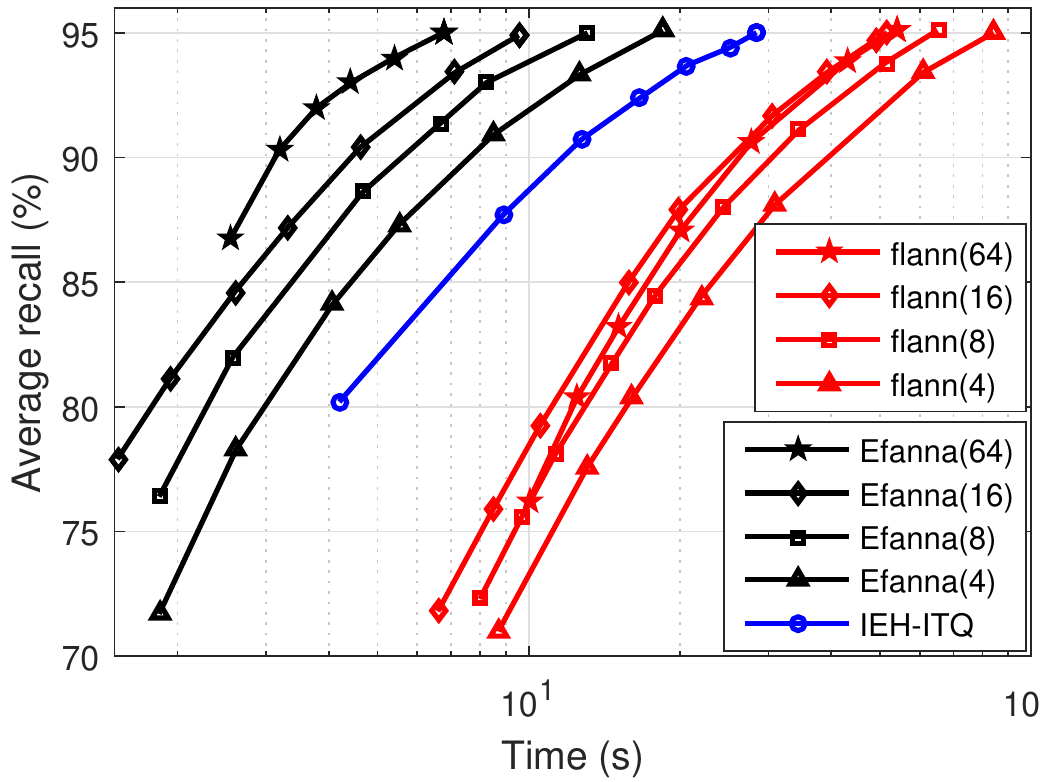}}
	\subfigure[SIFT1M 10NN]{\includegraphics[width=\DoubleFigureWidth]{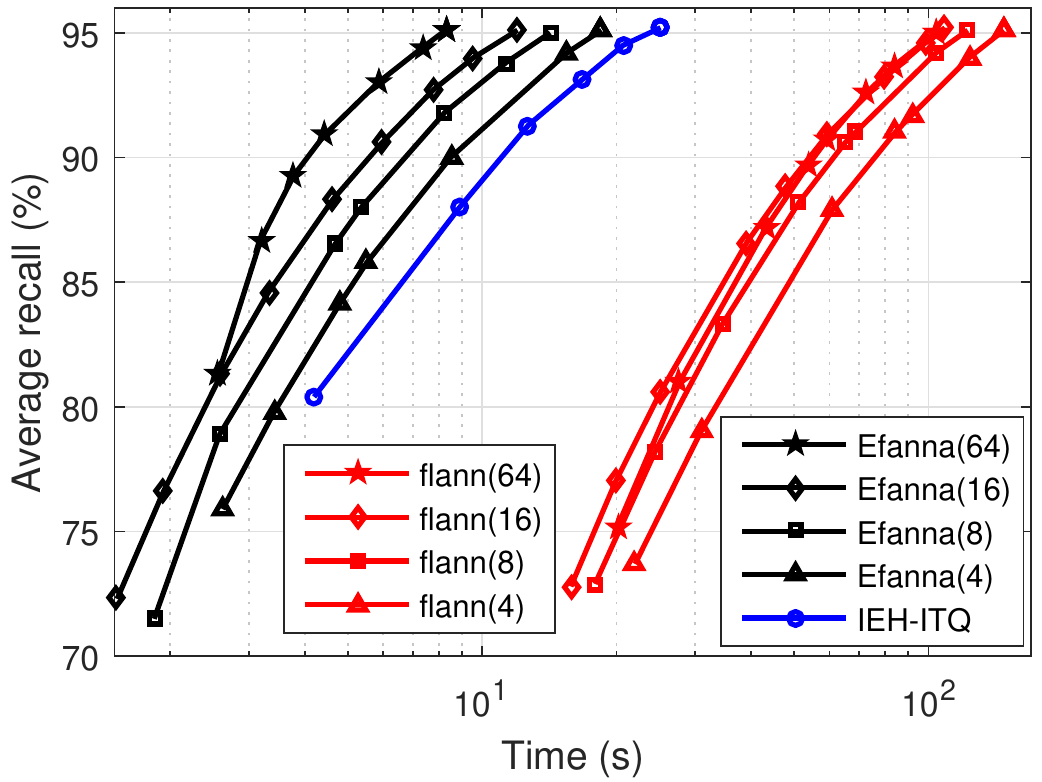}}
	\subfigure[SIFT1M 50NN]{\includegraphics[width=\DoubleFigureWidth]{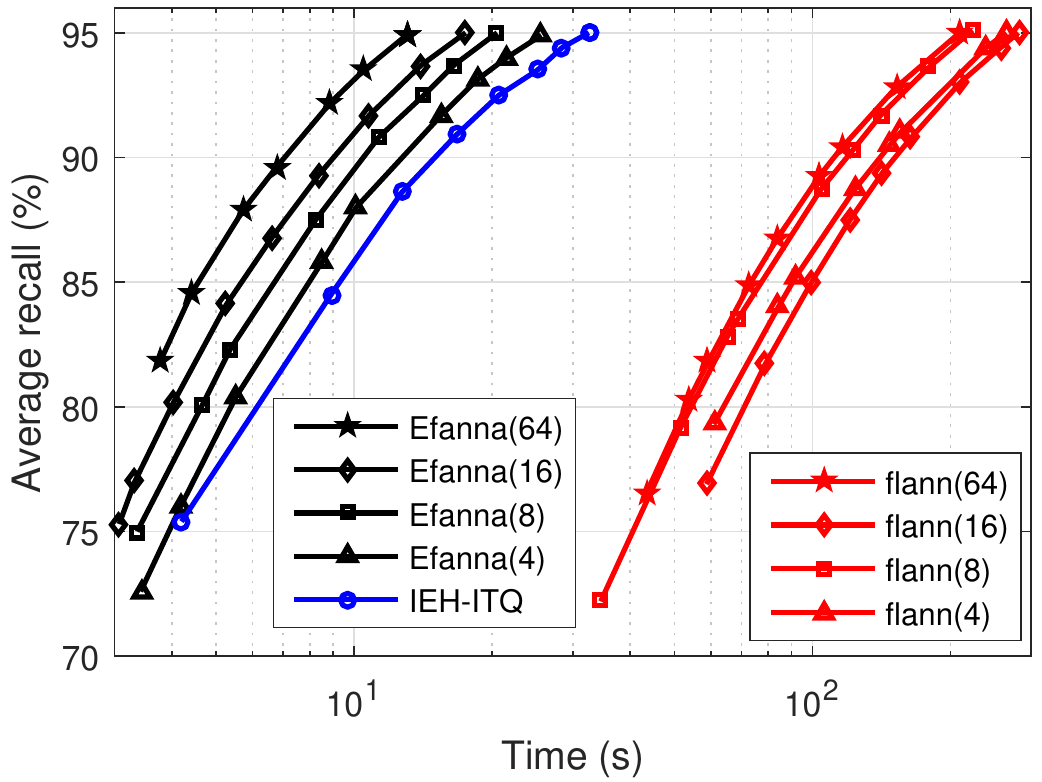}}
	\subfigure[SIFT1M 100NN]{\includegraphics[width=\DoubleFigureWidth]{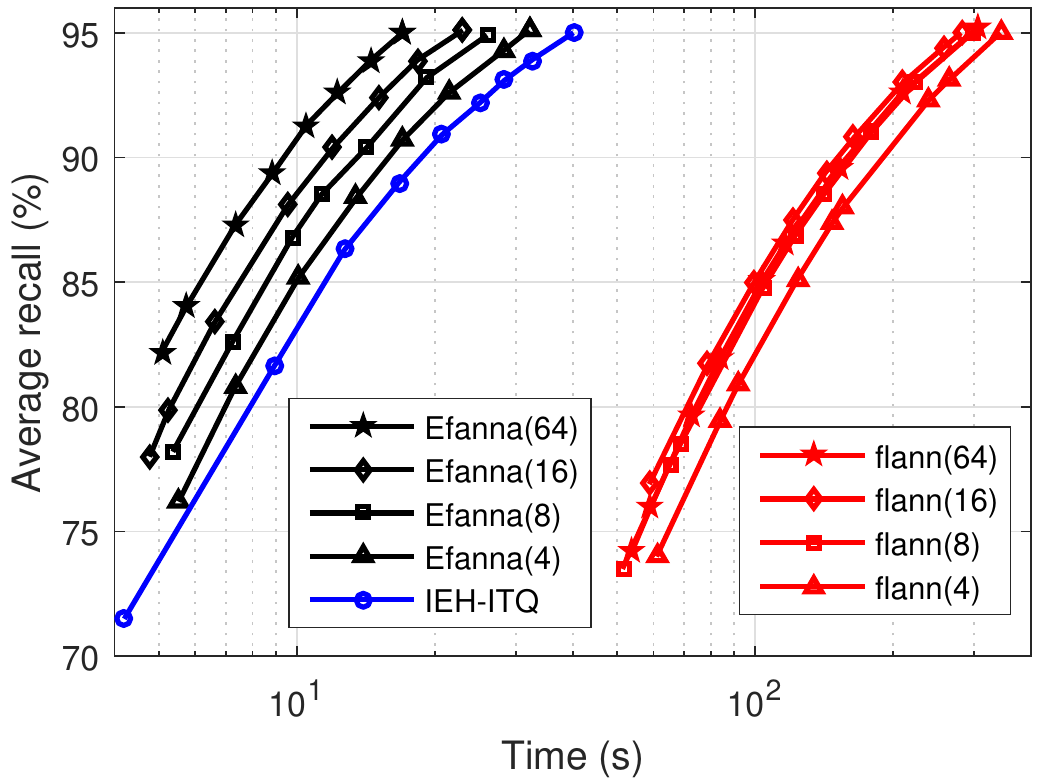}}
	\caption{Approximate nearest neighbor search results of 10,000 queries on SIFT1M. We use 4 $\sim$ 64 trees for EFANNA and flann-kdtrees respectively; The 10-NN ground truth graph is used for EFANNA and ITQ.}
	\label{SIFT_search_tree}
\end{figure*}

\begin{figure*}
	\centering
	\subfigure[GIST1M 1NN]{\includegraphics[width=\DoubleFigureWidth]{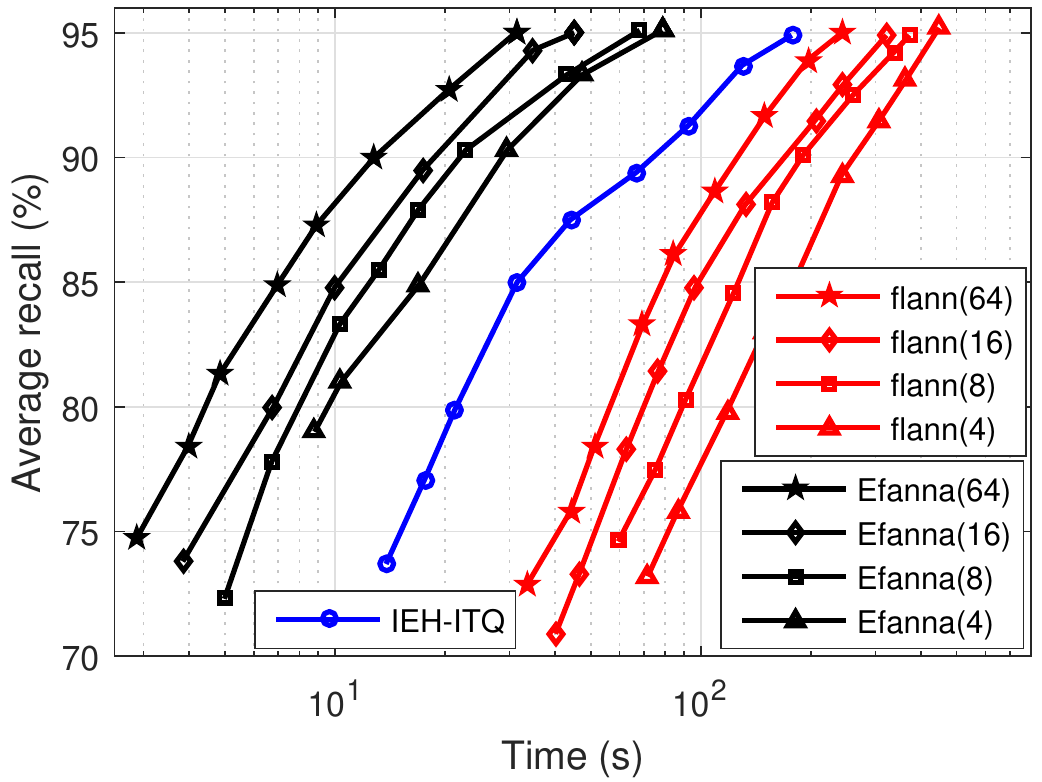}}
	\subfigure[GIST1M 10NN]{\includegraphics[width=\DoubleFigureWidth]{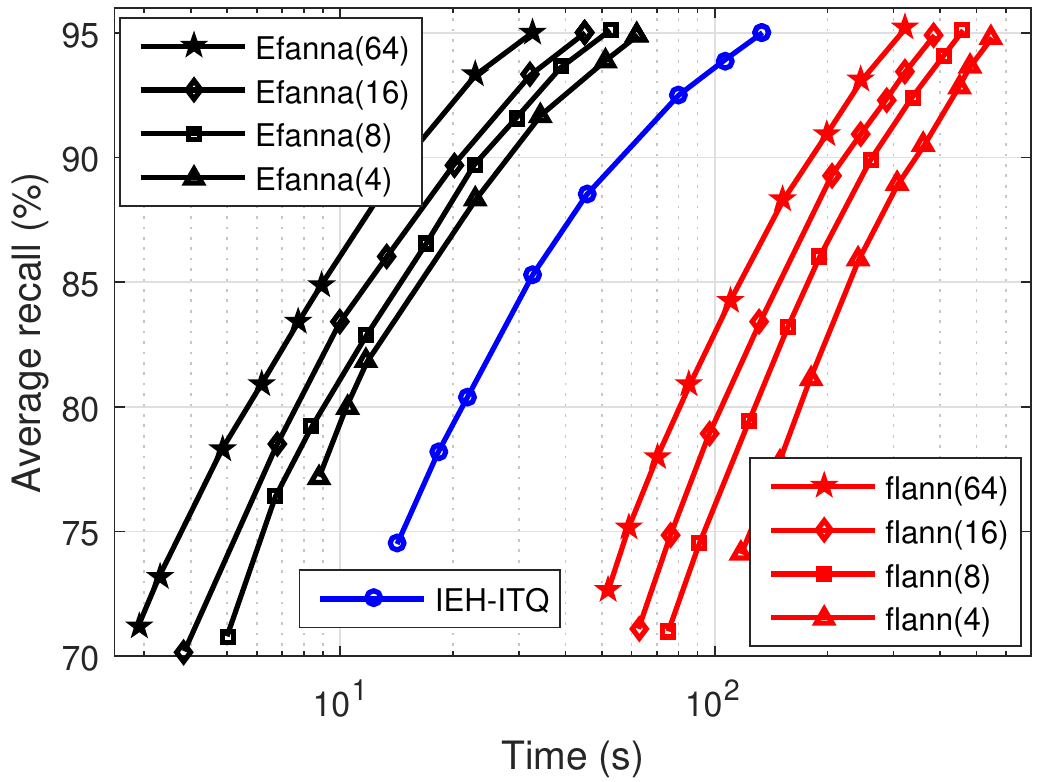}}
	\subfigure[GIST1M 50NN]{\includegraphics[width=\DoubleFigureWidth]{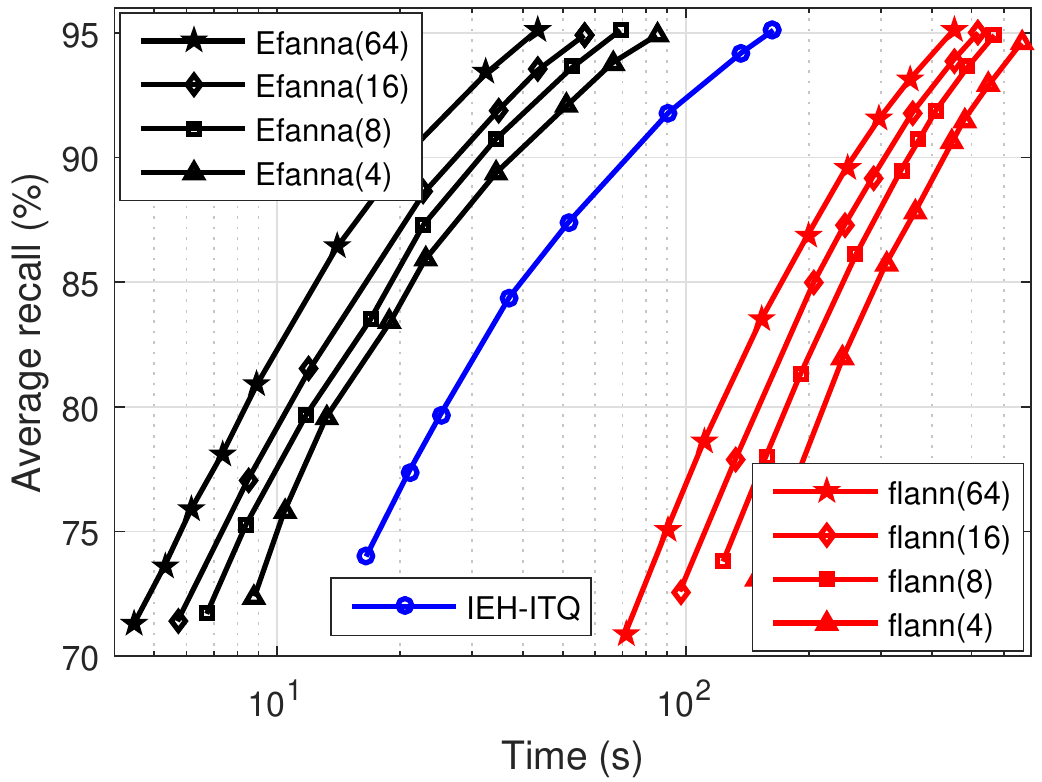}}
	\subfigure[GIST1M 100NN]{\includegraphics[width=\DoubleFigureWidth]{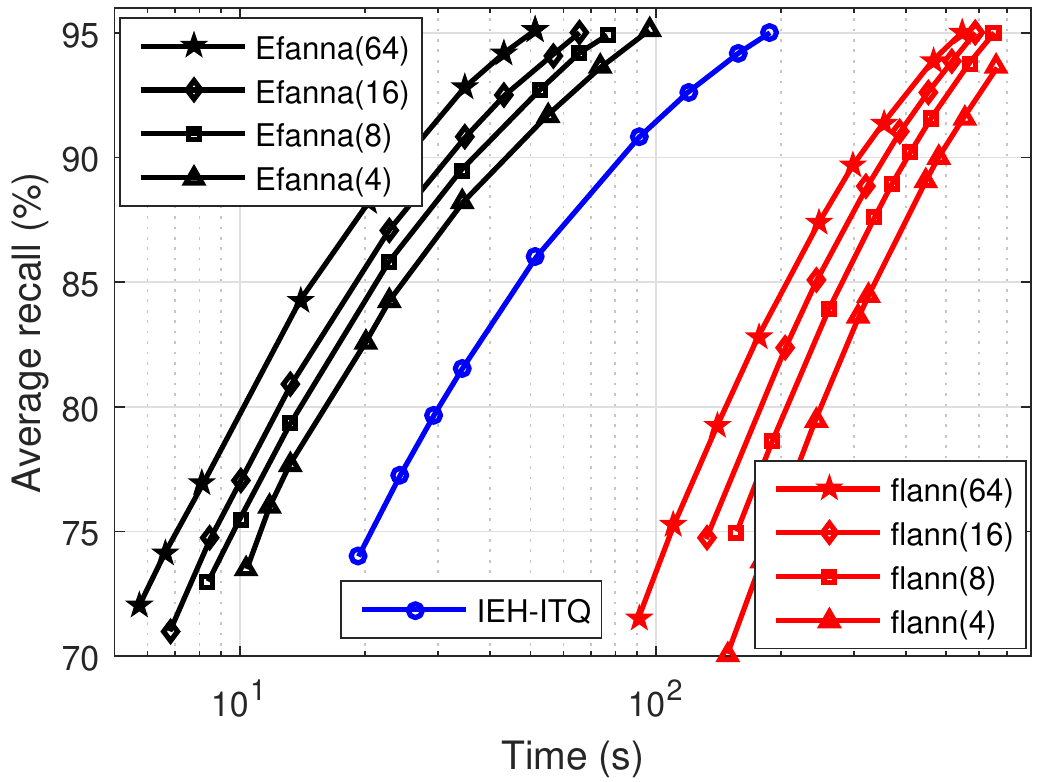}}
	\caption{Approximate nearest neighbor search results of 1,000 queries on GIST1M. We use 8 $\sim$ 64 trees for EFANNA and flann-kdtrees respectively; The 10-NN ground truth graph is used for EFANNA and ITQ.}
	\label{GIST_search_tree}
\end{figure*}

\begin{table*}[t]
	\caption{Index Size of Efanna and Flann with Different Number of Trees}
	\label{TreeIndexSizeTable}
	\centering
	\begin{tabular}{|c|c|c|c|c|c|c|c|c|}
		\hline
		\multirow{2}{*}{data set} & \multirow{2}{*}{algorithm}& \multicolumn{3}{c|}{ index size} & \multirow{2}{*}{algorithm}& \multicolumn{3}{c|}{ index size} \\
		\cline{3-5}
		\cline{7-9}
		& & tree & graph & all & & tree & graph & all\\
		\hline 
		\multirow{4}{1cm}{SIFT1M}  
		&flann(64-tree)& 3.85 GB & 0 & 3.85 GB &Efanna(64-tree, 10-NN)& 1.08 GB & 60.5 MB & 1.14 GB\\
		\cline{2-9}
		&flann(16-tree)& 997.5 MB & 0 & 997.5 MB &Efanna(16-tree, 10-NN)& 283.3 MB & 60.5 MB & 343.8 MB\\
		\cline{2-9}
		&flann(8-tree)& 506.7 MB & 0 & 506.7 MB &Efanna(8-tree, 10-NN)& 145.7 MB & 60.5 MB & 206.2 MB\\
		\cline{2-9}
		&flann(4-tree)& 261.2 MB & 0 & 261.2 MB &Efanna(4-tree, 10-NN)& 76.6 MB & 60.5 MB & 137.1 MB\\
		\hline
		\multirow{4}{1cm}{GIST1M}  
		&flann(64-tree)& 3.85 GB & 0 & 3.85 GB &Efanna(64-tree, 10-NN)& 1.08 GB & 60.5 MB & 1.14 GB\\
		\cline{2-9}
		&flann(16-tree)& 997.5 MB & 0 & 997.5 MB &Efanna(16-tree, 10-NN)& 283.3 MB & 60.5 MB & 343.8 MB\\
		\cline{2-9}
		&flann(8-tree)& 506.7 MB & 0 & 506.7 MB &Efanna(8-tree, 10-NN)& 145.7 MB & 60.5 MB & 206.2 MB\\
		\cline{2-9}
		&flann(4-tree)& 261.2 MB & 0 & 261.2 MB &Efanna(4-tree, 10-NN)& 76.6 MB & 60.5 MB & 137.1 MB\\
		\hline
	\end{tabular}
\end{table*}

\subsubsection{Results}

The time-accuracy curves of different algorithms on two data sets are shown in Fig. \ref{SIFT_graph} and Fig. \ref{GIST_graph} receptively. A number of interesting conclusions can be made. 

\begin{enumerate}
	\item EFANNA outperforms all the other algorithms on approximate $k$NN graph building. It can achieve more than 300 times speed-up over brute-force construction to reach 95\% accuracy. 
	Without parallelism, it takes a week to build a 10-NN graph on GIST1M using brute-force search. Now the time can be reduced to less than an hour by using EFANNA.
	
	\item We didn't get the source code of SGraph and FastKNN. So we implement their algorithms on our own. However, the performances shown in the two figures are quite different from what the original papers \cite{Gan2012Scalable} \cite{Zhang2013Fast} claim. One of the reasons may be the implementation. In the original FastKNN paper \cite{Zhang2013Fast}, the authors fail to add the hashing time into the total graph building time but actually should do. Fortunately, \cite{Gan2012Scalable} reported that SGraph achieved 100 times speed-up over brute-force on the SIFT1M at 95\% accuracy. And SGraph got 50 times speed-up over brute-force on the gist1M (384 dimensions) at 90\% accuracy. While EFANNA achieves over 300 times speed-up on both SIFT1M and GIST1M (960 dimensions). 
	
	\item LargeVis achieve significant better result than NN-expansion. However, NN-descent is better than LargeVis, especially when we want an accurate graph. This results confirm our assumption that many previous works had the misunderstanding of NN-descent. The result reported in their paper is actually NN-expansion \cite{Zhang2013Fast,tang2016visualizing} rather than NN-descent.  
	
	\item kGraph and NN-descent are actually the same algorithm. The only difference is that we implement NN-descent by ourselves and kGraph is an open library. The performance difference of these two methods should due to the implementation. 
	
	\item The only difference between EFANNA and NN-descent (kGraph) is the initialization. EFANNA uses randomized truncated KD-tree to build the initial graph while NN-descent (kGraph) use random initialization. 
	
	\item The performance advantage of EFANNA over NN-descent is larger on the SIFT1M than on the GIST1M. The reason maybe the GIST1M (960 dimensions) has higher dimensionality than the SIFT1M (128 dimensions). The KD-tree initialization becomes less effective when dimensions becomes high. The similar phenomena happens when we compare EFANNA and LargeVis. Since LargeVis uses random projection trees for initialization, this suggests random projection trees meybe better than KD-tree when the dimensions is high. Using random projection trees as the hierarchical structures of EFANNA can be the future work. 
	
\end{enumerate}

\begin{figure*}
	\centering
	\subfigure[SIFT1M 1NN]{\includegraphics[width=\DoubleFigureWidth]{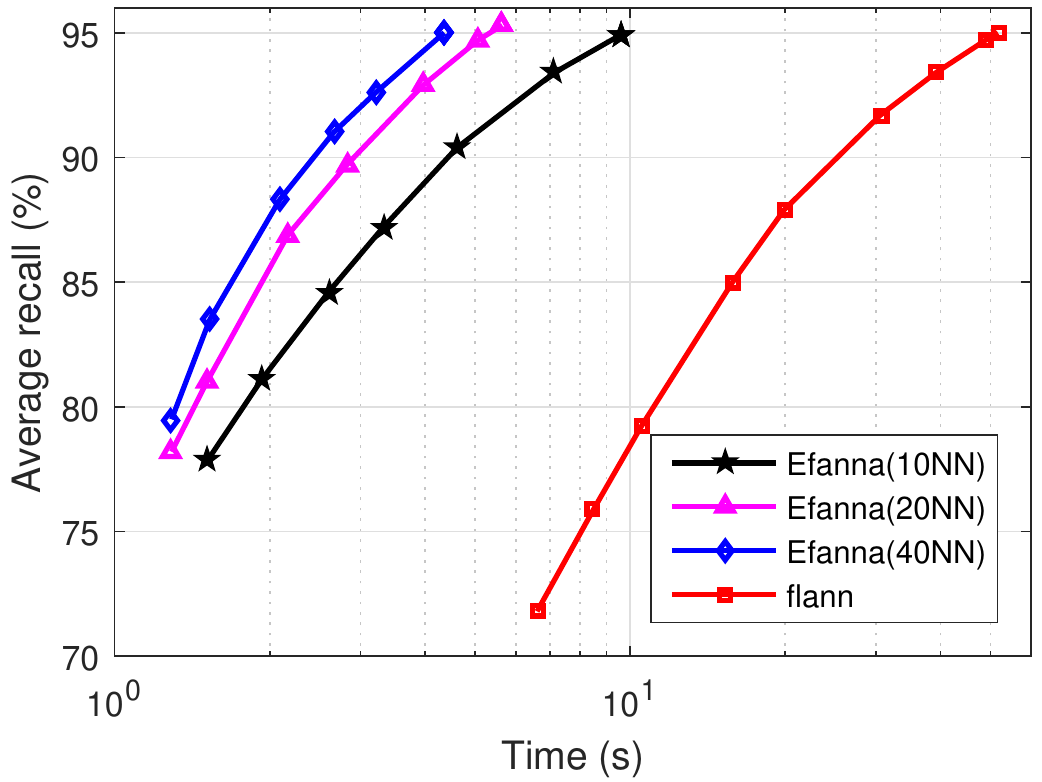}}
	\subfigure[SIFT1M 10NN]{\includegraphics[width=\DoubleFigureWidth]{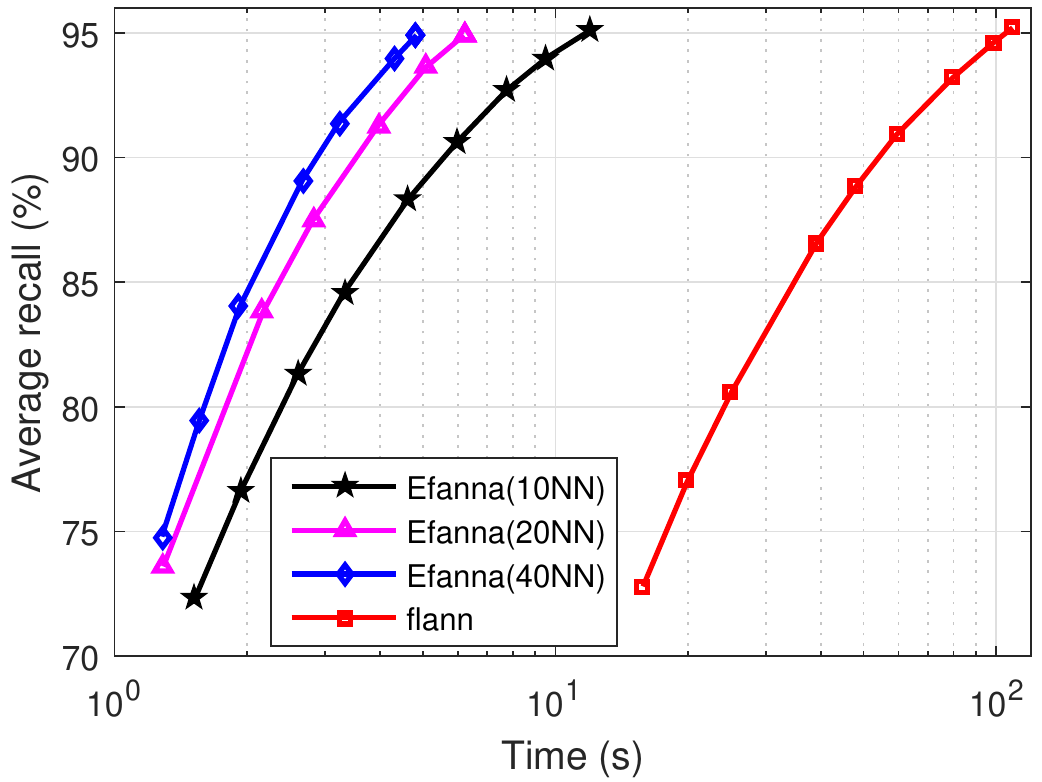}}
	\subfigure[SIFT1M 50NN]{\includegraphics[width=\DoubleFigureWidth]{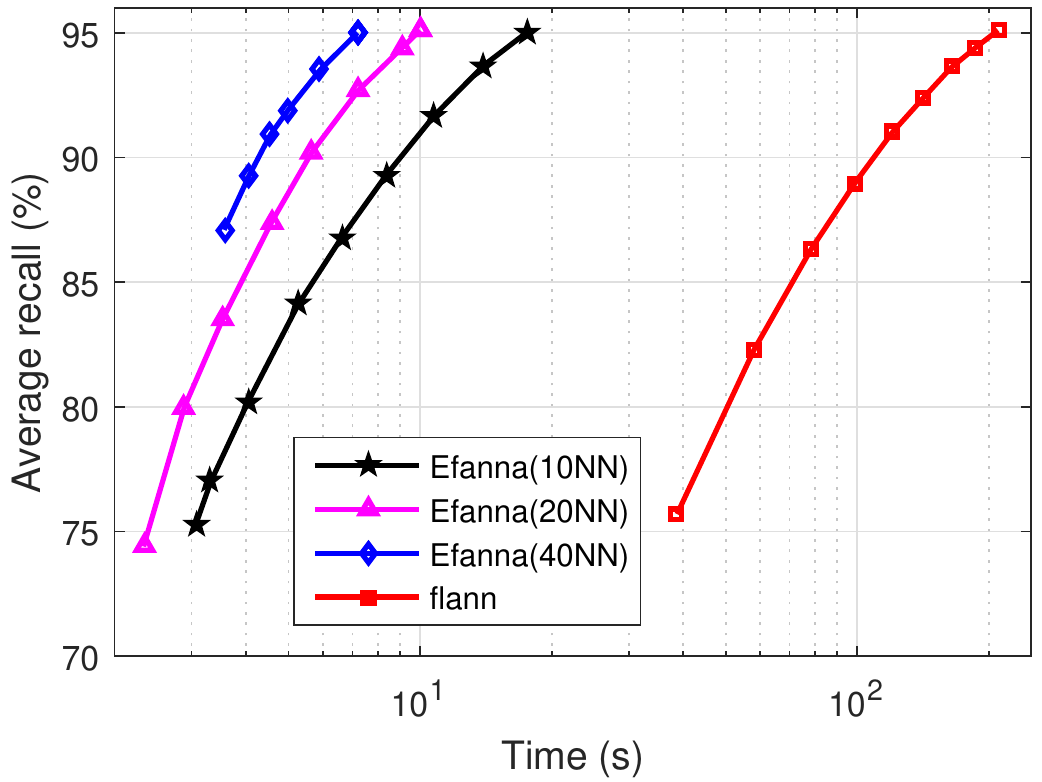}}
	\subfigure[SIFT1M 100NN]{\includegraphics[width=\DoubleFigureWidth]{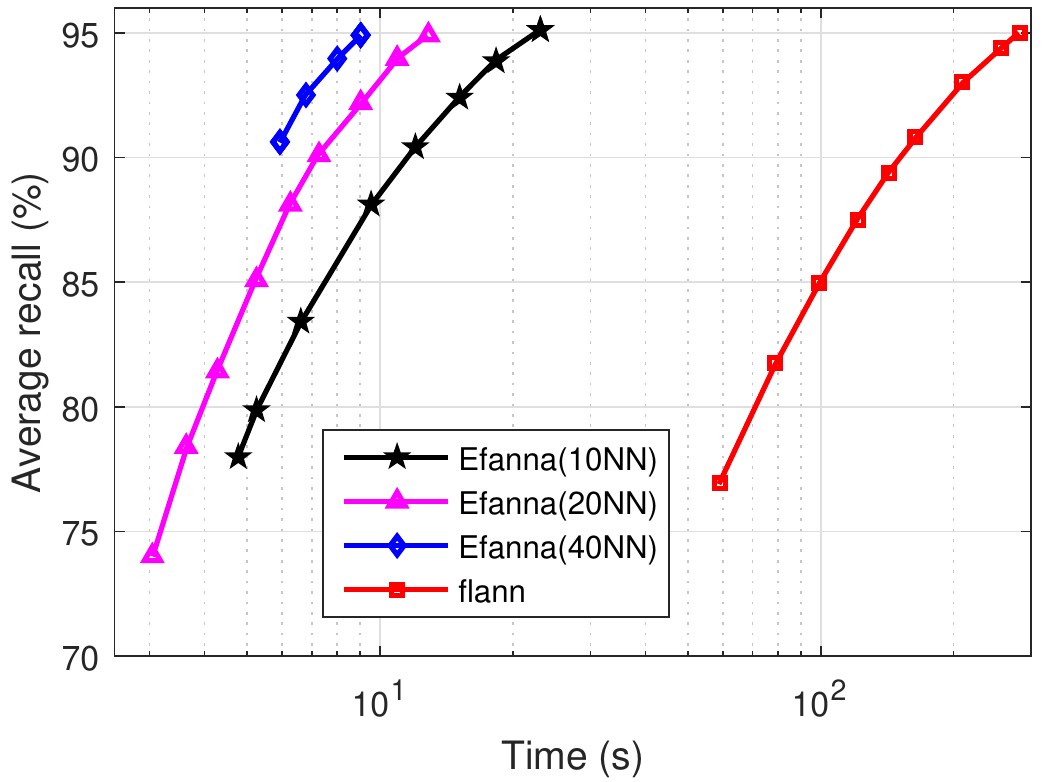}}
	\caption{Approximate nearest neighbor search results of 10,000 queries on SIFT1M. We use $k$NN graphs with various $k$ from 10 to 40 for for EFANNA respectively; Both EFANNA and flann-kdtrees use 16 trees.}
	\label{SIFT_search_graphW}
\end{figure*}

\begin{figure*}
	\centering
	\subfigure[GIST1M 1NN]{\includegraphics[width=\DoubleFigureWidth]{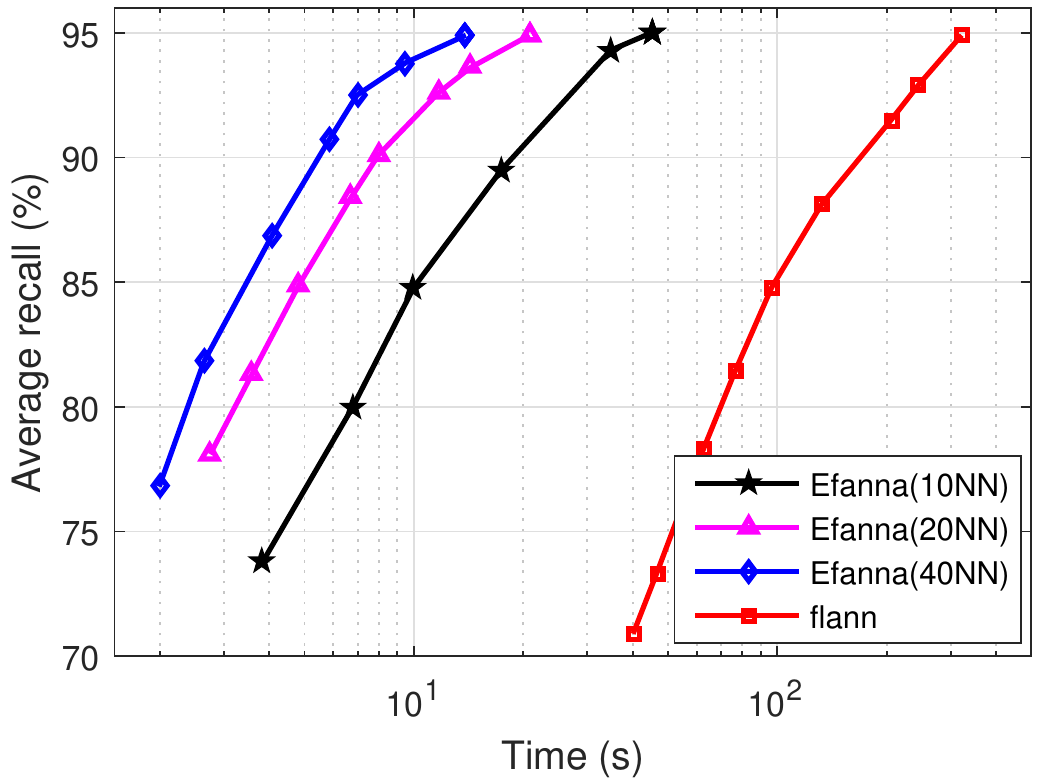}}
	\subfigure[GIST1M 10NN]{\includegraphics[width=\DoubleFigureWidth]{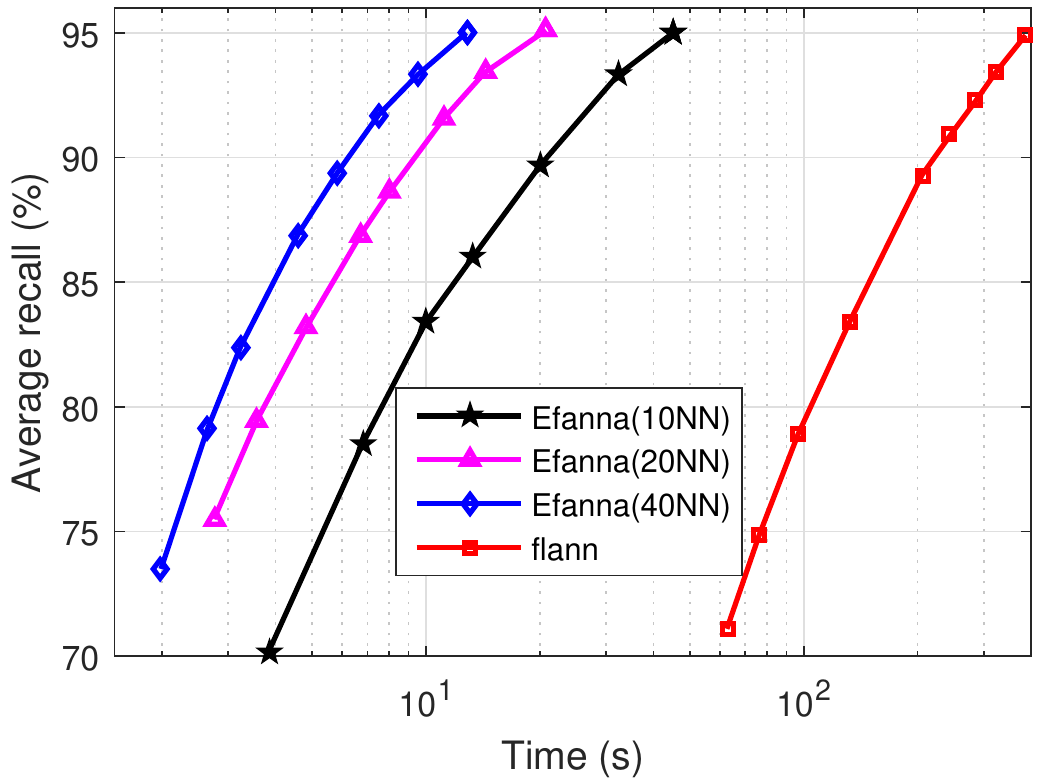}}
	\subfigure[GIST1M 50NN]{\includegraphics[width=\DoubleFigureWidth]{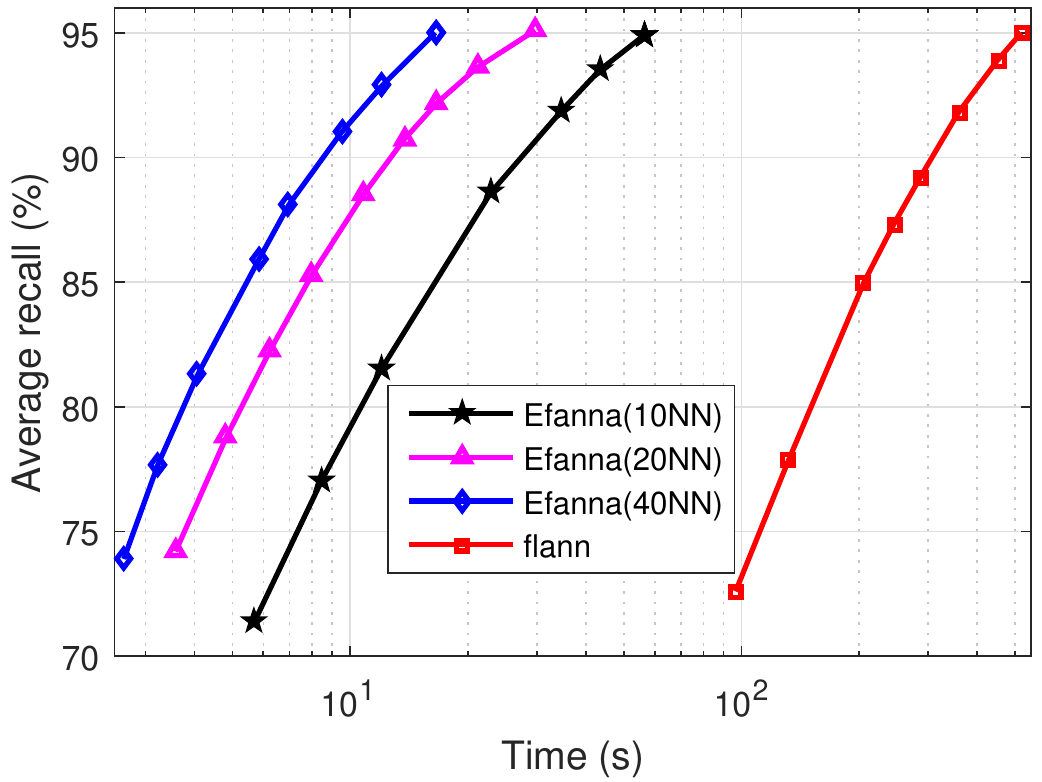}}
	\subfigure[GIST1M 100NN]{\includegraphics[width=\DoubleFigureWidth]{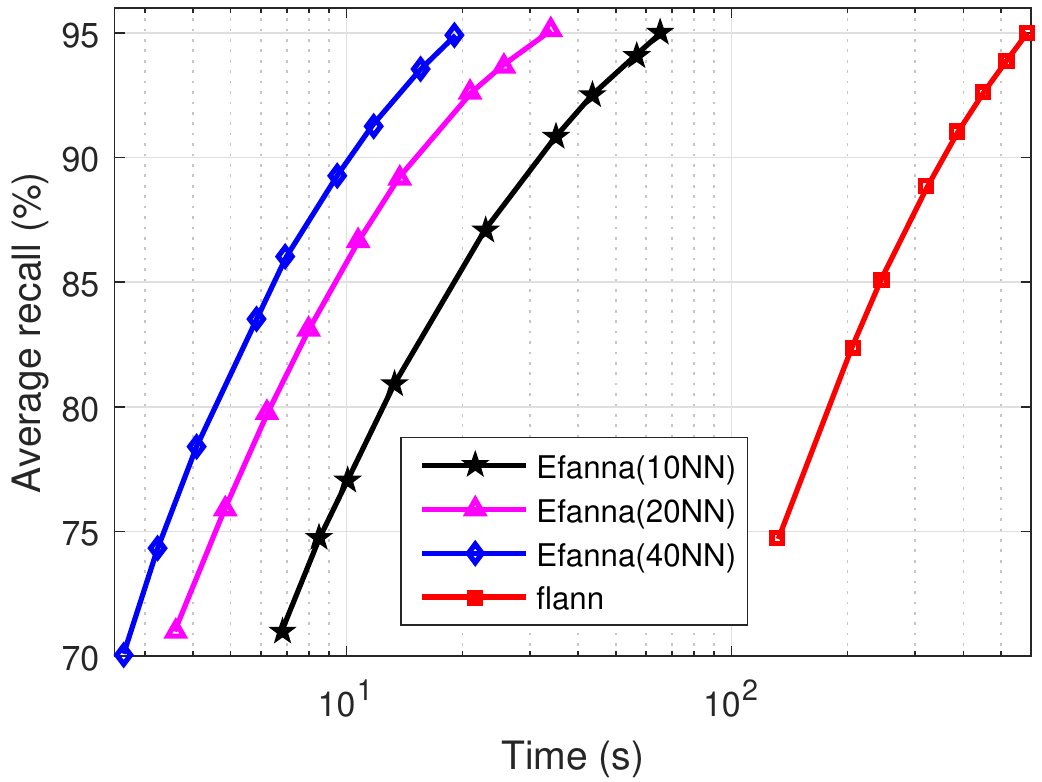}}
	\caption{Approximate nearest neighbor search results of 1,000 queries on GIST1M. We use $k$NN graphs with various $k$ from 10 to 40 for for EFANNA respectively; Both EFANNA and flann-kdtrees use 16 trees.}
	\label{GIST_search_graphW}
\end{figure*}

\begin{table*}[t]
	\caption{Index Size of Efanna with Different Number of $k$ for $k$NN Graph}
	\label{GraphIndexSizeTable}
	\centering
	\begin{tabular}{|c|c|c|c|c|}
		\hline
		\multirow{2}{*}{data set}  & \multirow{2}{*}{algorithm}& \multicolumn{3}{c|}{ index size} \\
		\cline{3-5}
		& & tree & graph & all \\
		\hline 
		\multirow{4}{1cm}{SIFT1M}  
		&Efanna(16-tree, 10-NN)& 283.3 MB & 60.5 MB & 343.8 MB\\
		\cline{2-5}
		&Efanna(16-tree, 20-NN)& 283.3 MB & 106.7 MB & 390.0 MB\\
		\cline{2-5}
		&Efanna(16-tree, 40-NN)& 283.3 MB & 182.8 MB & 446.1 MB\\
		\hline
		\multirow{4}{1cm}{GIST1M}  
		&Efanna(16-tree, 10-NN)& 283.3 MB & 60.5 MB & 343.8 MB\\
		\cline{2-5}
		&Efanna(16-tree, 20-NN)& 283.3 MB & 106.7 MB & 390.0 MB\\
		\cline{2-5}
		&Efanna(16-tree, 40-NN)& 283.3 MB & 182.8 MB & 446.1 MB\\
		\hline
	\end{tabular}
\end{table*}

\subsection{EFANNA with Approximate kNN Graphs}
\label{GraphSection}
The experimental results in the last section show that EFANNA can build an approximate $k$NN graph efficiently. However, there are no published results on the performance of graph based ANN search methods 
on an approximate $k$NN graph.

In this section, we evaluate the performance of EFANNA on approximate $k$NN graphs with various accuracy. The results on two data sets are shown in Fig .\ref{SIFT_search_graph} and \ref{GIST_search_graph} respectively.

From these two figures, we can see that the ANN search performance of EFANNA suffers from very little decrease in performance even when the graph is only ``half right''. Specifically, the ANN search preformance of EFANNA with a 60\% accurate 10-NN graph is still significant better than Flann-kdtree on SIFT1M. On GIST1M, EFANNA with a 57\% accurate 10-NN graph is significant better than Flann-kdtree. 

These results are significant because building a less accurate $k$NN graph using EFANNA is very efficient. Table \ref{IndexTimeTable} shows the indexing time of EFANNA and Flann-kdtree. If a 60\% accurate graph is used, the indexing time of EFANNA is similar to that of Flann-kdtree. Combing the results in Table \ref{IndexSizeTable}, we can see that comparing with Flann-kdtree, EFANNA takes similar indexing time, smaller index size and significant better ANN search performance.

Why EFANNA can get such a good ANN search performance even with a ``half right'' graph? Table \ref{GraphAccuracyTable} may explain the reason. The accuracy defined in Eqn. \ref{eq:accuracy} uses the size of $R$ and $R'$. The former is the true nearest neighbors set while the latter is the returned nearest neighbors set of an algorithm. In the previous experiments, we fix the sizes of both $R$ and $R'$ as 10. Table \ref{GraphAccuracyTable} reports the results by varying the size of $R$ form 10 to 100. We cam see that a 60\% accurate 10-NN graph constructed by EFANNA in SIFT1M means 60\% of all the neighbors are true 10-nearest neighbors. And the remaining 40\% neighbors are not randomly select from the whole dataset. Actually, 98.9\% of the neighbors are true 100-nearest neighbors. These results show that the approximate $k$NN graphs constructed by EFANNA are very good approximation of the ground truth $k$NN graph.

\begin{figure*}
	\centering
	\subfigure[SIFT1M 1NN]{\includegraphics[width=\DoubleFigureWidth]{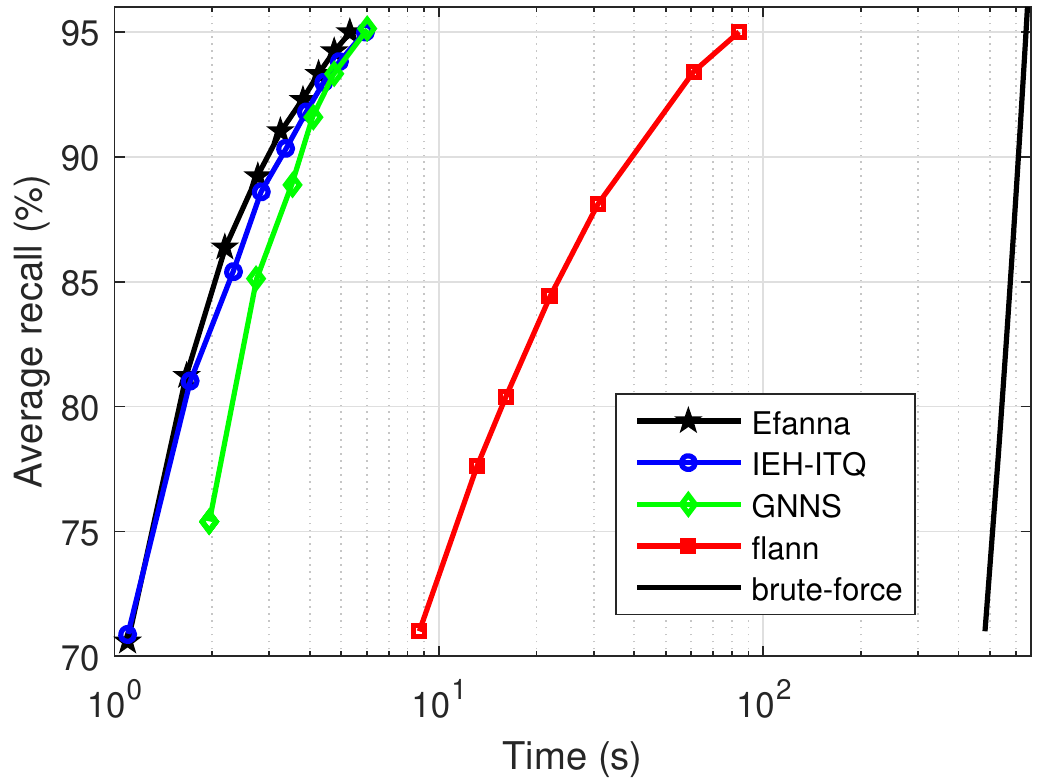}}
	\subfigure[SIFT1M 10NN]{\includegraphics[width=\DoubleFigureWidth]{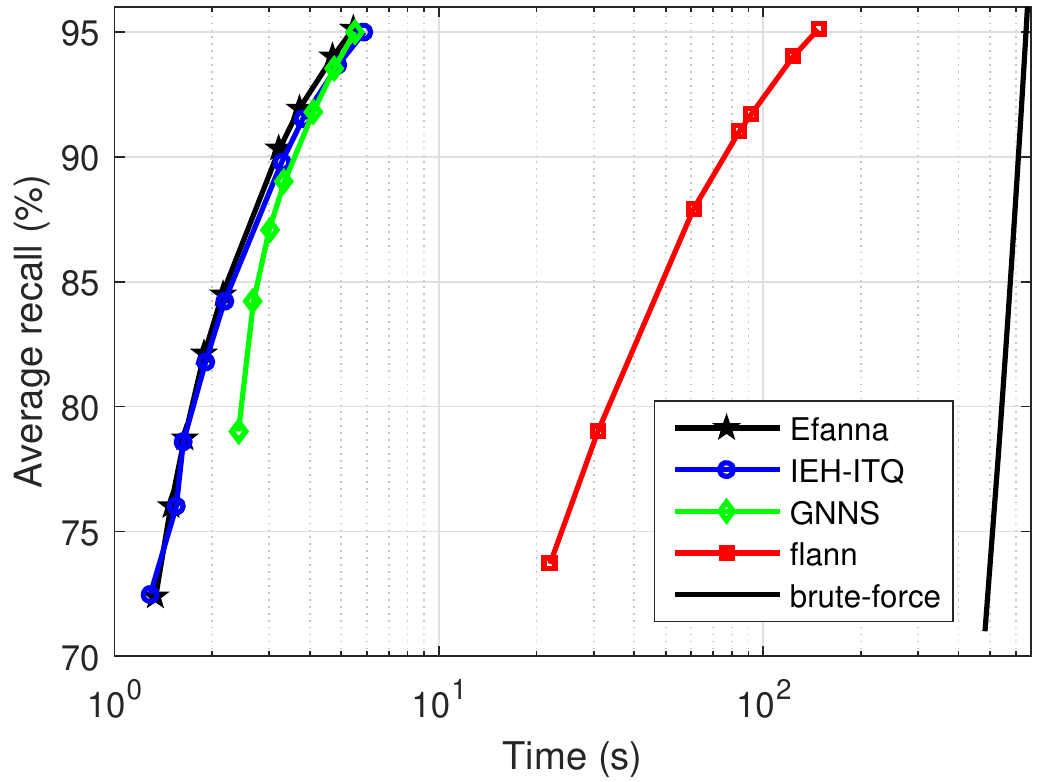}}
	\subfigure[SIFT1M 50NN]{\includegraphics[width=\DoubleFigureWidth]{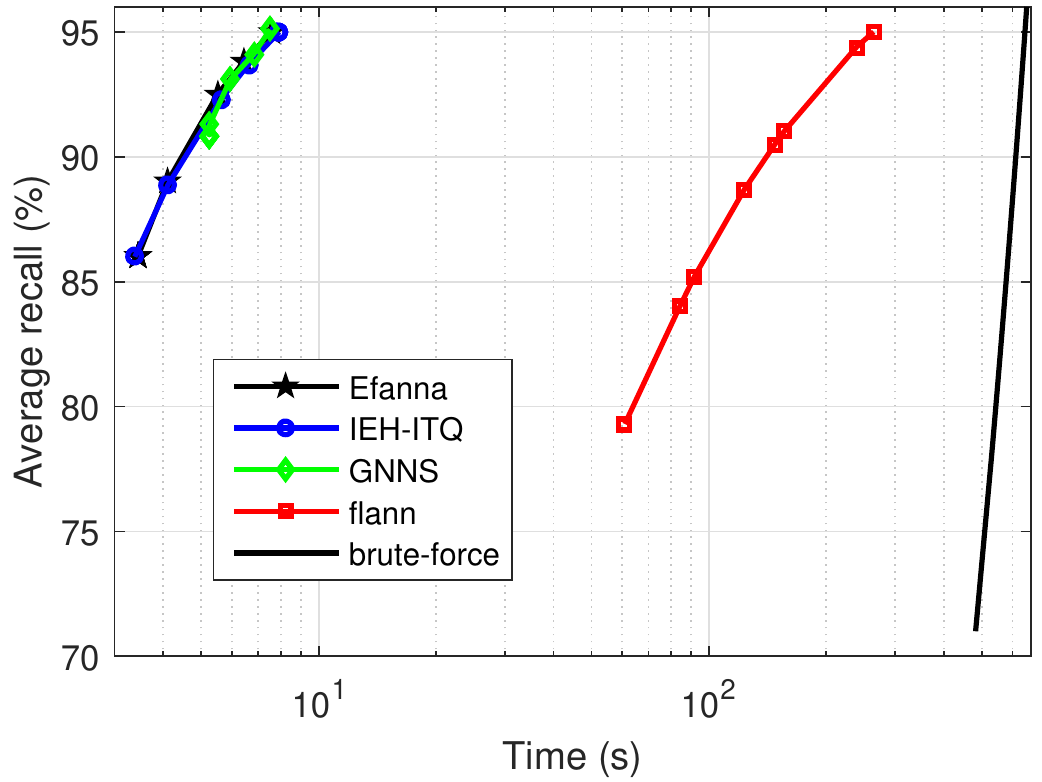}}
	\subfigure[SIFT1M 100NN]{\includegraphics[width=\DoubleFigureWidth]{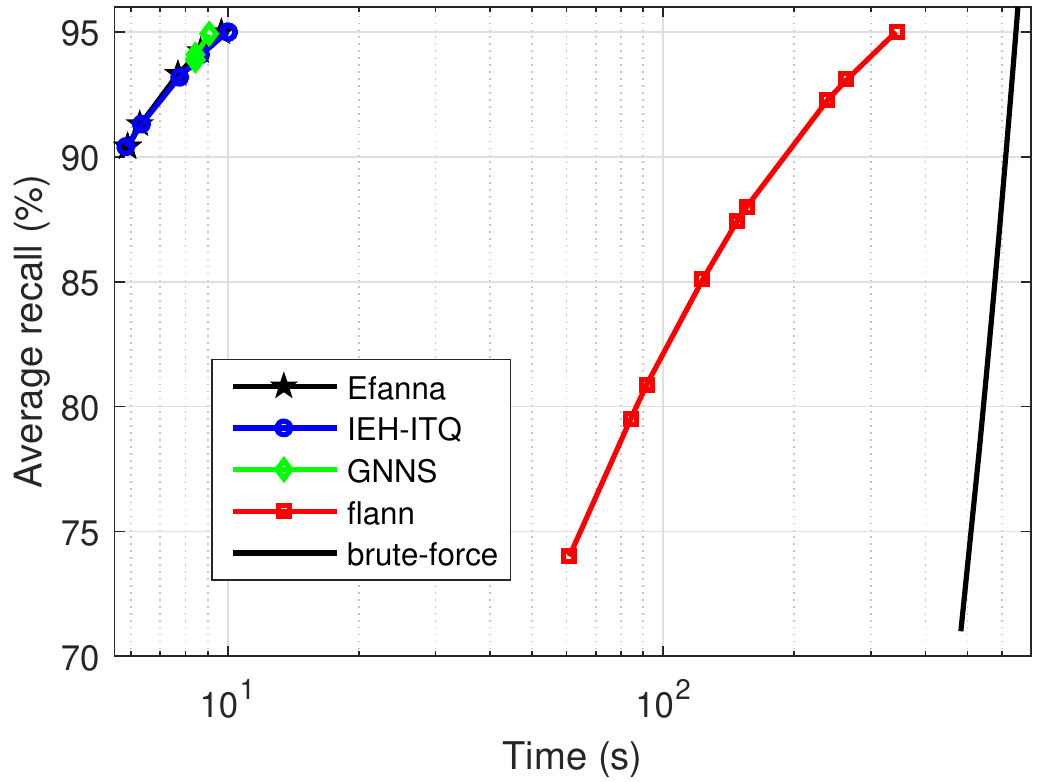}}\\
	\subfigure[SIFT1M 1NN]{\includegraphics[width=\QuaterFigureWidth]{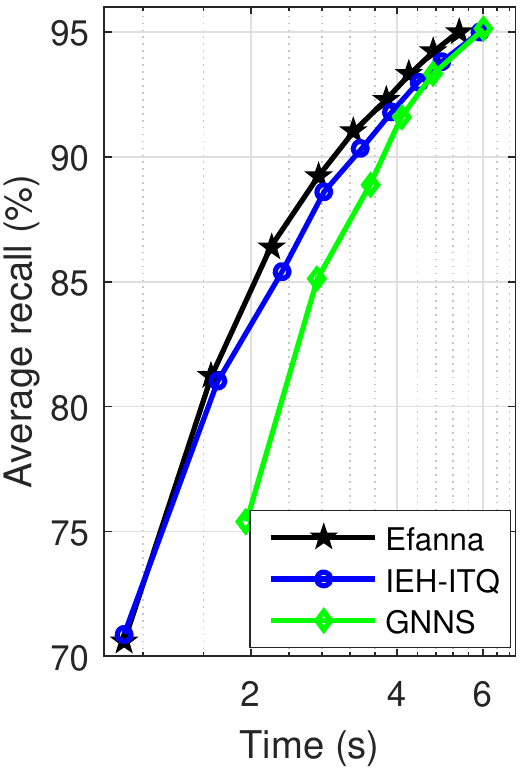}}
	\subfigure[SIFT1M 10NN]{\includegraphics[width=\QuaterFigureWidth]{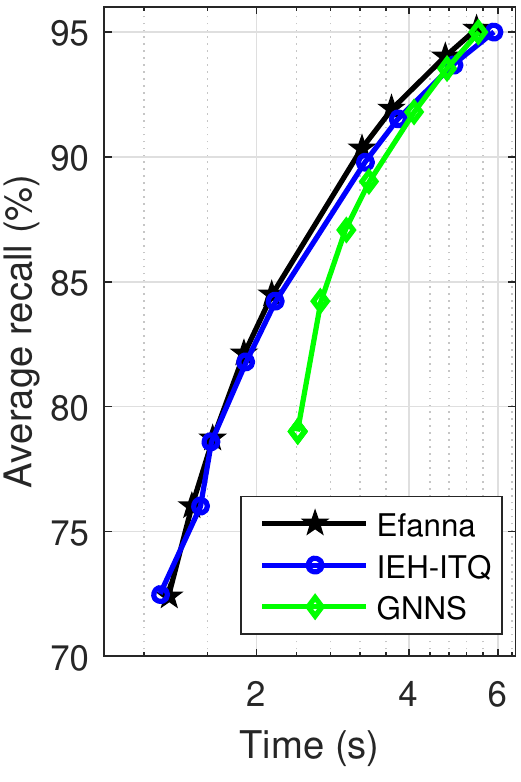}}
	\subfigure[SIFT1M 50NN]{\includegraphics[width=\QuaterFigureWidth]{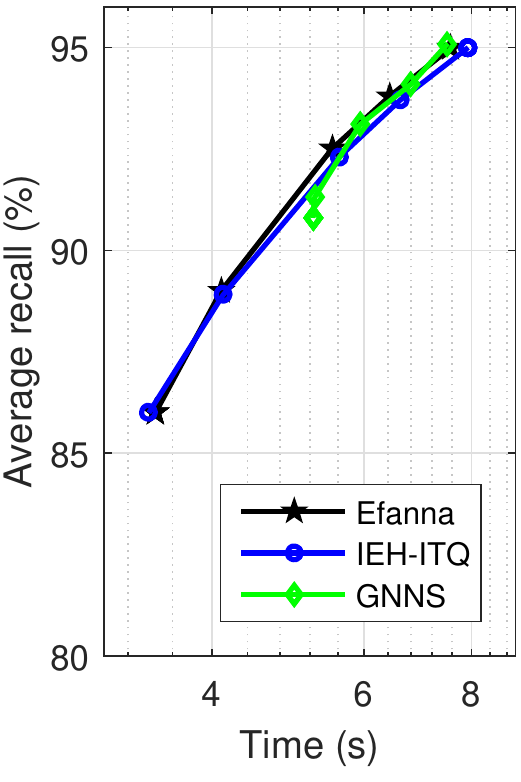}}
	\subfigure[SIFT1M 100NN]{\includegraphics[width=\QuaterFigureWidth]{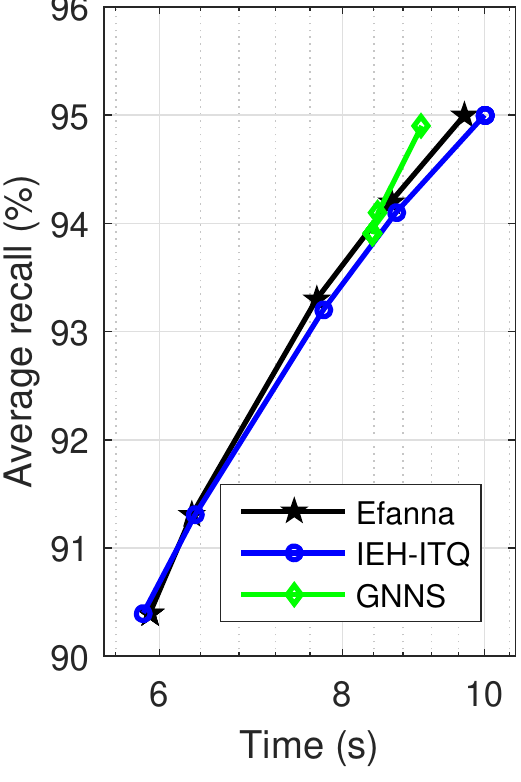}}
	\caption{ANN search results of 10,000 queries on SIFT1M. All the four ANN search methods used the same index size as shown in the table (\ref{SameIndexSizeTable}).}
	\label{SIFT_search_mem}
\end{figure*}

\begin{figure*}
	\centering
	\subfigure[GIST1M 1NN]{\includegraphics[width=\DoubleFigureWidth]{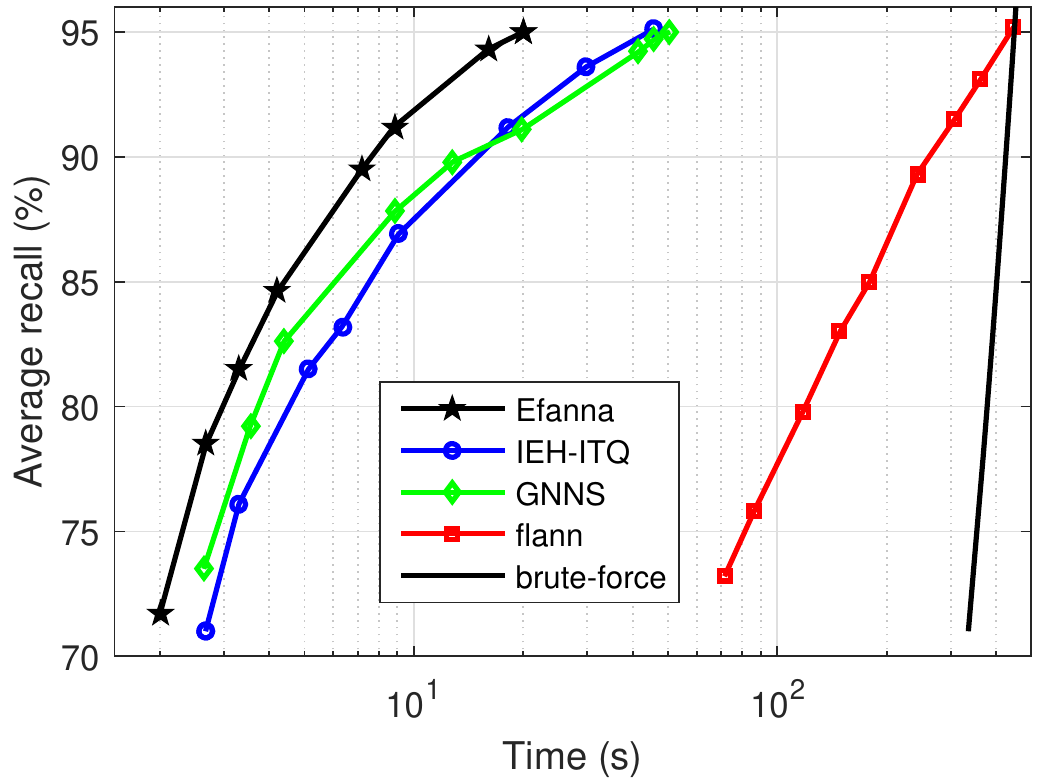}}
	\subfigure[GIST1M 10NN]{\includegraphics[width=\DoubleFigureWidth]{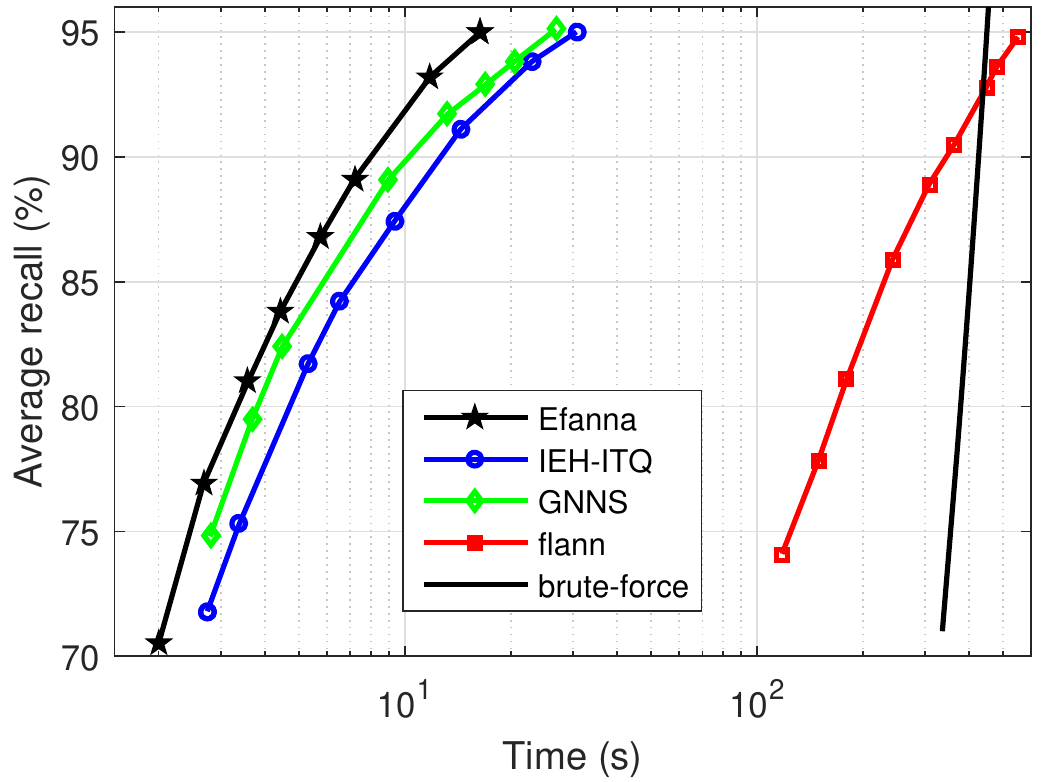}}
	\subfigure[GIST1M 50NN]{\includegraphics[width=\DoubleFigureWidth]{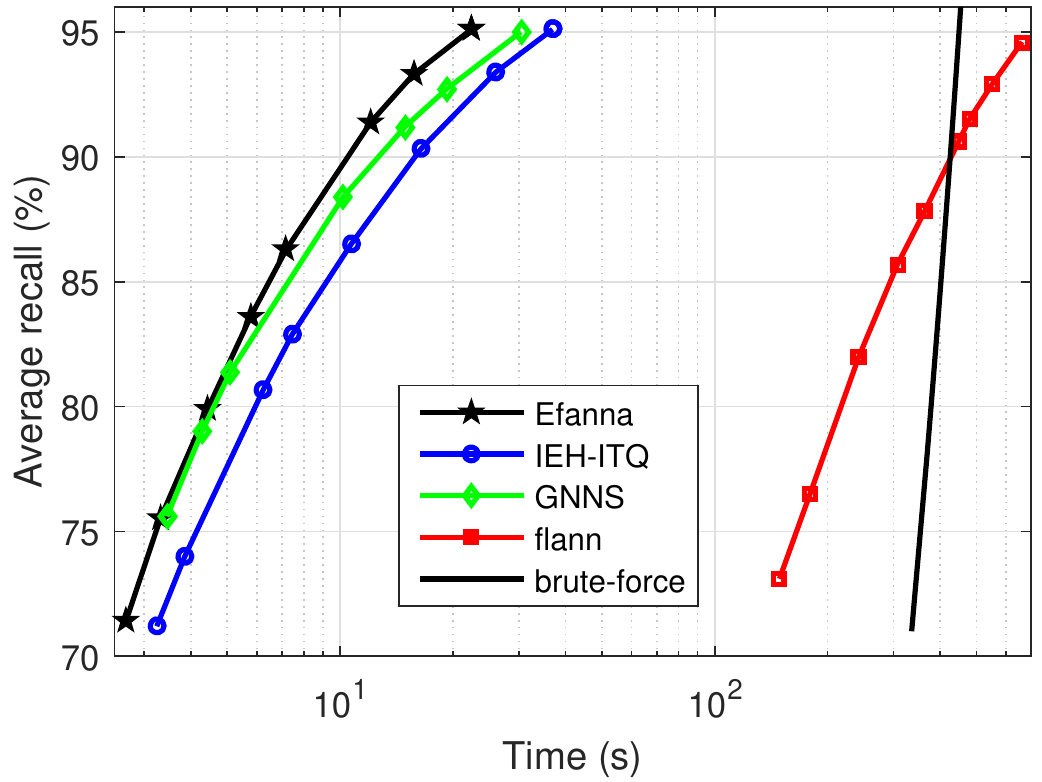}}
	\subfigure[GIST1M 100NN]{\includegraphics[width=\DoubleFigureWidth]{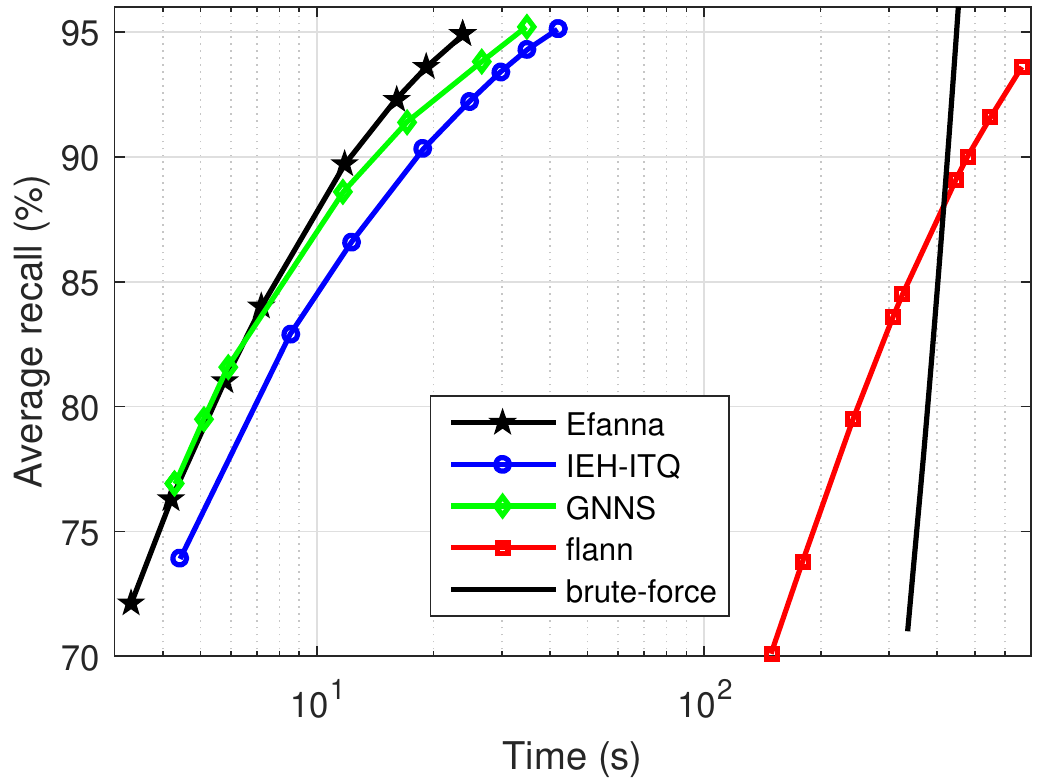}}
	\caption{ANN search results of 1,000 queries on GIST1M. All the four ANN search methods used the same index size as shown in the table (\ref{SameIndexSizeTable}).}
	\label{GIST_search_mem}
\end{figure*}

\begin{table*}[t]
	\caption{Index Size of Different Algorithms}
	\label{SameIndexSizeTable}
	\centering
	\begin{tabular}{|c|c|c|c|c|}
		\hline
		\multirow{2}{*}{data set} & \multirow{2}{*}{algorithms}& \multicolumn{3}{c|}{ index size} \\
		\cline{3-5}
		& & tree (hash table) & graph & all \\
		\hline 
		\multirow{6}{1cm}{SIFT1M}  
		&flann(4-tree)& 261.2 MB & 0 & 261.2 MB\\
		\cline{2-5}
		&Efanna(4-tree, 40-NN)& 76.6 MB & 182.8 MB & 259.4 MB\\
		\cline{2-5}
		&GNNS(60-NN)& 0 & 266.7 MB & 266.7 MB \\
		\cline{2-5}
		&IEH-ITQ (32 bit, 40-NN)& 82.7 MB & 182.8 MB & 265.5 MB \\
		\hline
		\multirow{6}{1cm}{GIST1M}  
		&flann(4-tree)& 261.2 MB & 0 & 261.2 MB\\
		\cline{2-5}
		&Efanna(4-tree, 40-NN)& 76.6 MB & 182.8 MB & 259.4 MB\\
		\cline{2-5}
		&GNNS(60-NN)& 0 & 266.7 MB & 266.7 MB \\
		\cline{2-5}
		&IEH-ITQ (32 bit, 40-NN)& 82.7 MB & 182.8 MB & 265.5 MB \\
		\hline
		\multicolumn{5}{|l|}{The index size here is the size in the memory, not the size on the disk.}\\
		\hline
	\end{tabular}
\end{table*}

\subsection{EFANNA with Different Number of Trees}
\label{TreeSection}

In the previous experiments, EFANNA uses 16 truncated kd-trees for search initialization. Table \ref{IndexSizeTable} shows that these trees consume a large number of  memory space. In this experiment, we want to explore how the number of trees will influence the performance of EFANNA on ANN search. Throughout this experiment, we use the 10-NN ground truth graph.

The ANN search results on SIFT1M and GIST1M are shown in Fig. \ref{SIFT_search_tree} and FIg. \ref{GIST_search_tree} respectively. We simply compare with IEH-ITQ and FLANN, because IEH-ITQ is the second best algorithm on ANN search in our previous experiment while FLANN also has the tree number parameter.

From Fig. \ref{SIFT_search_tree} and \ref{GIST_search_tree}, we can see that with less number of trees, the ANN search performances of both EFANNA and Flann decrease. However, with only 4 trees, EFANNA still significantly better than IEH-ITQ (especially on the GIST1M data set). While the index sizes can be significantly reduced as suggested by Table \ref{TreeIndexSizeTable}. With 4 trees, the index size of EFANNA is smaller than that of IEH-ITQ.

The results in this section show the flexibility of EFANNA over other graph based ANN search methods. One can easily make trade-off between index size and search performance.

\subsection{EFANNA with Different Number of $k$ in $k$NN Graph}
\label{kGraphSection}

The EFANNA index contains two parts: the truncated kd-trees and the $k$NN graph. If we regard the $k$NN graph as an $N \times k$ matrix, we can use the ``width'' of the graph to denote $k$. In the previous section, we have checked the performance of EFANNA with different number of trees. Now we will show how the ``width'' of $k$NN graph influences ANNS performance of EFANNA.

Fig.\ref{SIFT_search_graphW} and \ref{GIST_search_graphW} show the ANNS performance of EFANNA with graph 10NN, 20NN, 40NN on SIFT1M and GIST1M. The index size are showed in TABLE \ref{GraphIndexSizeTable} respectively. From TABLE \ref{GraphIndexSizeTable} we can see that, from 10NN to 40NN, the size of EFANNA index grows gradually. Besides, in Fig.\ref{SIFT_search_graphW}, \ref{GIST_search_graphW}, the performance of increase with the growing of graph `width'. 

Compared with Fig. \ref{SIFT_search_tree}, \ref{GIST_search_tree}, we can get a conclusion that widening the graph provides more boost on ANNS performance of EFANNA than add more trees. And from the comparison between TABLE \ref{TreeIndexSizeTable} and \ref{GraphIndexSizeTable}, we find that with equal extra memory cost, widening graph is a better choice then using more trees. 

However, we should also notice that the performance boost does not increase linearly with the `width' of the graph. In other words, there may exists an upper bound of performance boost by increasing EFANNA index size, either from the aspect of tree or graph.

\subsection{ANN Search Comparison with Same Index Size}
\label{SameIndexMem}
The results in previous section suggest the comparisons in section \ref{simple_search_compare} is not quite fair due to different index size of different algorithms. In this section, we try to compare different algorithms with (almost) equal index size. 

We reported the performance of EFANNA, IEH-ITQ, GNNS and flann's KD-tree. We do not compare with IEH-LSH simply because IEH-ITQ is better than IEH-LSH. We do not compare with kGraph because GNNS is almost identical with kGraph. 

We restrict the index size of each algorithm to about 265 MB. Finally, we use 4 trees for flann's KD-tree; 4 trees and 40NN graph for EFANNA; 1 table and 40NN graph for IEH-ITQ; 60NN graph for GNNS. See TABLE \ref{SameIndexSizeTable} for details on how we organize the index to get almost equal size. Fig. \ref{SIFT_search_mem}, \ref{GIST_search_mem} show the performance of these algorithms on SIFT1M and GIST1M. 

On both two datasets, graph based methods achieve over 20x speed up over flann's KD-tree with the same index size. Particularly, EFANNA is about 30x faster than flann's KD-tree. This suggests the advantage of graph based methods over traditional tree structure based methods.

Compared with the results in Fig. \ref{SIFT_search_gt} and \ref{GIST_search_gt}, we can find that the performance gain achieved by EFANNA over IEH-ITQ and GNNS (kGraph) becomes smaller as the ``width'' of graph grows. This indicates the impact of good initialization for NN-expansion becomes small as the ``width'' of graph grows.

With the same index size, EFANNA and IEH-ITQ still have small advantage than GNNS on SIFT1M when the recall is low. At a high recall level (\eg, 95\%), the performances of three algorithms are almost the same. Particularly, when we search for 100NN, the performance of GNNS (random initialization) is better than EFANNA and IEH-ITQ at 95\% recall level. This is actually expected because good initializations require additional time. If the information provided by the $k$NN graph is enough, random initialization is the best choice.

On GIST1M, EFANNA still have the advantage over IEH-ITQ and GNNS (kGraph), which again suggest that GIST1M is a ``harder'' dataset for ANNS problem. We surprisingly find that GNNS is better than IEH-ITQ which suggests truncated KD-tree (used in EFANNA) is a better choice than hashing (ITQ) used for initialization. It's interesting to investigate better initialization algorithms.

\section{The EFANNA Library}\label{sec5}

The work in this paper is released as an open source library. Please access the code at Github\footnote{https://github.com/fc731097343/efanna}.

\section{Conclusion}\label{sec6}

The goal of this research is to provide a fast solution, EFANNA, for both ANN search and approximate $k$NN graph construction problems. On ANN search, we use hierarchical structures to provide better initialization for NN-expansion. And on graph construction, we use a divide-and-conquer way to construct an initial graph and refine it with NN-descent. Extensive experiments shows that EFANNA outperforms previous algorithms significantly both in approximate $k$NN graph construction and ANN search. Meanwhile, EFANNA also shows great flexibility for different scenarios.

\ifCLASSOPTIONcompsoc
\section*{Acknowledgments}
\else
\section*{Acknowledgment}
\fi
We acknowledge Xiuye Gu (gxy0922@zju.edu.cn) for implementing the FastKNN algorithm and Xiaoshuang Zhang (zxs19930207@126.com) for implementing the SGraph algorithm. This work was supported in part by National Basic Research Program of China (973 Program) under Grant 2013CB336500 and National Youth Top-notch Talent Support Program.
Any opinions, findings, and conclusions expressed here are those of the
authors and do not necessarily reflect the views of the funding agencies.

\bibliographystyle{ieee}
\bibliography{Efanna}%

\end{document}